\documentclass[preprint,12pt]{elsarticle}

\usepackage{amssymb}
\usepackage{amsmath}
\usepackage{amsthm}
\usepackage{booktabs}
\usepackage{threeparttable}
\usepackage{subcaption} 

\usepackage[utf8]{inputenc} 
\usepackage[T1]{fontenc}    
\usepackage[colorlinks=true,allcolors=blue]{hyperref}       
\usepackage{url}            
\usepackage{booktabs}       
\usepackage{amsfonts}       
\usepackage{nicefrac}       
\usepackage{microtype}      
\usepackage{xcolor}         

\usepackage{graphicx} 
\usepackage{amsmath} 
\usepackage{algorithm}
\usepackage{algpseudocode}
\usepackage{float}

\usepackage[numbers]{natbib}

\newcommand{\R}{\mathbb{R}}
\newcommand{\vc}{\mathbf}

\newcommand{\eps}{\varepsilon}
\newcommand{\id}{\mathrm{d}}

\newtheorem{prop}{Proposition}

\journal{Computer Methods in Applied Mechanics and Engineering}

\begin{document}

\begin{frontmatter}



\title{mLaSDI: Multi-stage latent space dynamics identification}


\author[llnl]{William Anderson \corref{correspondingauthor}}
\author[llnl]{Seung Whan Chung}
\author[llnl]{Robert Stephany}
\author[llnl]{Youngsoo Choi}
\cortext[correspondingauthor]{Corresponding Author: \href{mailto:anderson316@llnl.gov}{\texttt{anderson316@llnl.gov}}}

\address[llnl]{Lawrence Livermore National Laboratory, Livermore, CA 94550, United States}
\begin{abstract}
Accurately solving partial differential equations (PDEs) is essential across many scientific disciplines. 
However, high-fidelity solvers can be computationally prohibitive, motivating the development of reduced-order models (ROMs).
Recently, Latent Space Dynamics Identification (LaSDI) was proposed as a data-driven, non-intrusive ROM framework. 
LaSDI compresses the training data via an autoencoder and learns user-specified ordinary differential equations (ODEs) governing the latent dynamics, enabling rapid predictions for unseen parameters.
While LaSDI has produced effective ROMs for numerous problems, the autoencoder must simultaneously reconstruct the training data and satisfy the imposed latent dynamics. These are often competing objectives which limit accuracy, particularly for complex or high-frequency phenomena.
To address this limitation, we propose multi-stage Latent Space Dynamics Identification (mLaSDI).
With mLaSDI, we train LaSDI sequentially in stages. 
After training the initial autoencoder, we train additional decoders which map the latent trajectories to residuals from previous stages.
This staged residual learning, combined with periodic activation functions, enables recovery of high-frequency content without sacrificing interpretability of the latent dynamics.
{We further provide an error decomposition separating autoencoder and latent dynamics contributions, and prove that additional training stages cannot increase the training residual. }
Numerical experiments on a multiscale oscillating system, unsteady wake flow, and the 1D-1V Vlasov equation demonstrate that mLaSDI achieves significantly lower reconstruction and prediction errors, while requiring less training time and reduced hyperparameter tuning compared to standard LaSDI.
\end{abstract}

\begin{graphicalabstract}
\\
\\
\includegraphics[width=\linewidth]{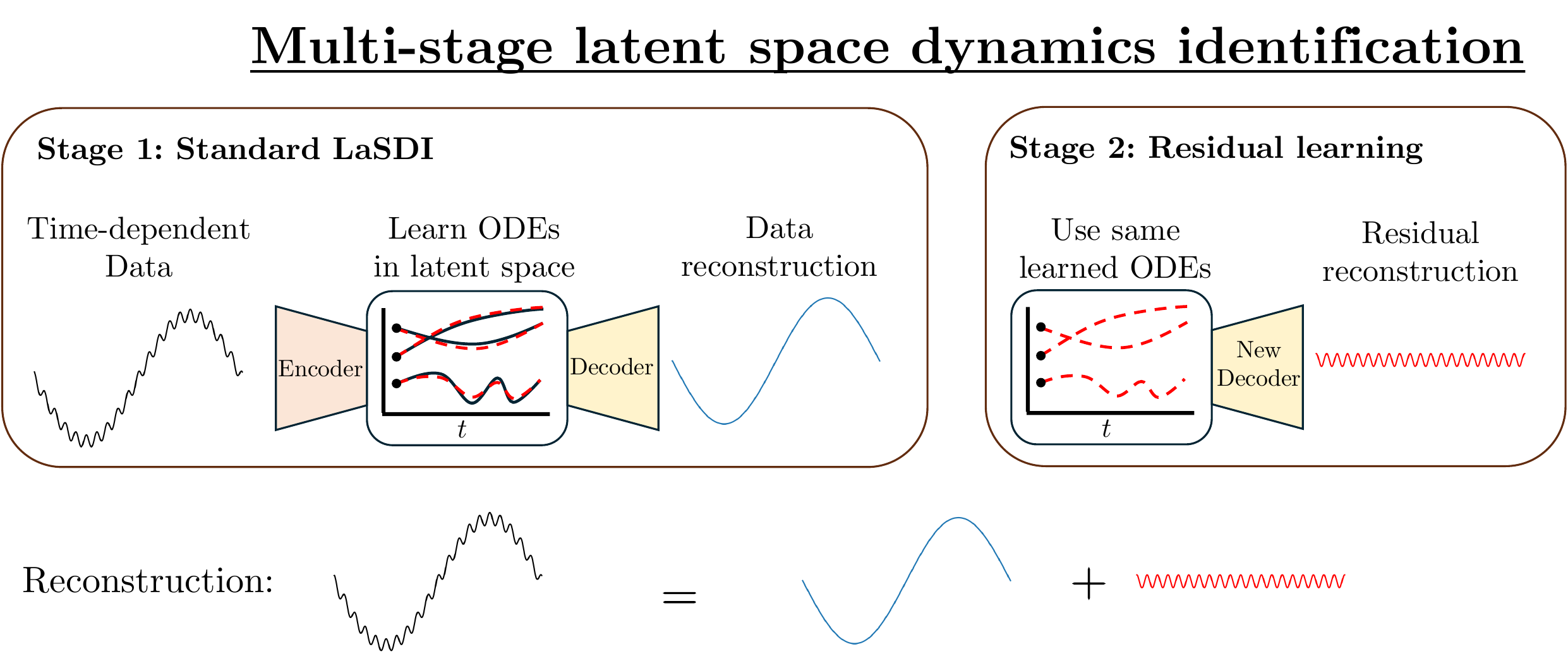}
\end{graphicalabstract}

\begin{highlights}
\item Residual learning enables accurate reduced-order models with interpretable latent space dynamics
\item Multi-stage training reduces hyperparameter sensitivity for neural networks
\item High-frequency dynamics are recovered through staged residual correction with periodic activation functions
\end{highlights}

\begin{keyword}



Data-driven modeling \sep reduced-order modeling \sep latent space dynamics learning \sep autoencoder

\end{keyword}

\end{frontmatter}



\section{Introduction}

Advances in computational power and numerical techniques have enabled increasingly accurate and complex simulations of time-dependent partial differential equations (PDEs).
High-fidelity numerical simulations improve understanding in many scientific fields such as engineering \cite{Calder2018,Cummings2015,Jones2020}, physics \cite{Thijssen2007,Vasileska2017} and biology \cite{Noble2002}.
While accurate, these simulations are often computationally prohibitive, which necessitates the use of reduced-order models (ROMs). 
In this work, we focus on parametric PDEs, where input parameters influence the initial conditions or underlying physics.

There exists rich theory for projection-based ROMs, where the governing equations are known \cite{Benner2015,Berkooz1993}. 
However, linear projection-based approaches such as the proper orthogonal decomposition struggle with advection-dominated problems \cite{Patera2007,Safonov1989}.
Nonlinear projection techniques have shown improved accuracy for these problems \cite{Bonneville2024a,Diaz2024,Kim2022,choi2021space,choi2020gradient,mcbane2021component,copeland2022reduced,choi2020sns,kim2021efficient,chung2024train}, but remain intrusive,  requiring knowledge of the governing equations.

Another approach is to use non-intrusive reduced-order models, which are purely data-driven.
These approaches develop ROMs without any knowledge of the underlying PDE.
Yet, many non-intrusive approaches lack interpretability, which makes it difficult to generalize results beyond the training regime.
This is the main motivation for the reduced-order modeling framework introduced by \citet{Fries2022}, Latent Space Dynamics Identification (LaSDI). 
LaSDI trains an autoencoder and learns interpretable dynamics of the compressed data by applying Sparse Identification of Nonlinear Dynamics (SINDy) \cite{Brunton2016} in the latent space.
By interpolating the learned ODE coefficients, LaSDI enables rapid predictions for unseen parameter values.
Since its inception, there have been several improvements and variations to the LaSDI algorithm \cite{Bonneville2024,He2023,He2025,Park2024,Tran2024,he2025thermodynamically,chung2025latent,he2022certified,brown2023data}. 
For example, developments include simultaneous training of the autoencoder and SINDy with residual-based active learning \cite{He2023}, the introduction of Gaussian Processes (GPs) to interpolate for SINDy coefficients \cite{Bonneville2024}, implementations of weak-form SINDy to deal with noisy data \cite{He2025, Tran2024}, and implementing physical constraints in the latent space \cite{Park2024,he2025thermodynamically}.

Despite these advances, a fundamental tension remains: the autoencoder must simultaneously reconstruct training data accurately and produce latent trajectories that satisfy the prescribed ODE structure.
These competing objectives may compromise both reconstruction quality and predictive accuracy, particularly for complex or high-frequency dynamics.
There is a well-known spectral bias of neural networks toward low frequencies \cite{rahaman19a,Wang2024,xu2020}, which is exacerbated by imposing smooth latent dynamics on our autoencoder.
In practice, achieving acceptable performance with LaSDI may require large autoencoders and extensive hyperparameter tuning.


To overcome these limitations, we introduce \textbf{multistage Latent Space Dynamics Identification (mLaSDI)}, a framework that extends any variant of LaSDI by training multiple networks sequentially.
After the initial autoencoder is trained, subsequent stages introduce new decoders which map the same latent trajectories to the residuals from previous stages.
Because all stages share a common latent representation, we preserve interpretability of the latent dynamics while dramatically improving reconstruction accuracy. We will show that for some examples, mLaSDI improves reconstruction accuracy by an order of magnitude with reduced training time and less hyperparameter sensitivity.
\noindent Source code for this project is based on Gaussian Process-based LaSDI (GPLaSDI): \hyperlink{https://github.com/LLNL/GPLaSDI}{https://github.com/LLNL/GPLaSDI}.

\subsection{Related Works} 
Several works have applied equation-learning algorithms \cite{Brunton2016, Schmidt2009} to approximate the underlying latent space dynamics \cite{Benner2020, Cranmer2020, Issan2022, Qian2020}. 
Perhaps the most closely related work to LaSDI is that of \citet{Champion2019}, who train an autoencoder to learn a SINDy representation in the latent space.
However, their method is not parameterized and therefore less generalizable than LaSDI.

Multiple authors have also proposed the idea of training sequential neural networks in other contexts to achieve increased accuracy.
{
The authors in \cite{kani2017} perform intrusive multi-stage model by combining the proper orthogonal decomposition and Discrete Empirical Interpolation Method with stacked recurrent neural networks to achieve more efficient model order reduction.
For non-intrusive methods, b}oth \cite{Aldirany2024} and \cite{Wang2024} introduce multi-stage neural networks, where at each stage a new network is introduced to learn the error from the previous stages. 
In \cite{Wang2024} the initial weights of the networks are altered in order to learn the residual more quickly, while \cite{Aldirany2024} increases the network size at each stage. 
For physics-informed neural networks, \cite{Howard2025} proposes a stacked architecture combining linear and nonlinear networks at each stage.
Several authors \cite{Vincent2010,Yu2022, Zabalza2016} have also explored the use of stacked autoencoders to increase reconstruction accuracy.

While multi-stage architectures exist in other contexts, none of these methods address the interpretability--accuracy trade-off inherent to LaSDI and other models with an interpretable latent space.
To our knowledge, mLaSDI is the first multi-stage architecture designed for interpretable latent dynamics, where each stage must respect dynamical constraints.
This is a notable departure from previous residual-learning networks and a fundamentally different challenge compared to prior stacked architectures.

We emphasize that the goal of this manuscript is not to provide comparison with other popular residual-based learning approaches, but instead to provide a framework that can fundamentally improve results for many of the different LaSDI approaches.
We demonstrate the performance of mLaSDI on several benchmark problems, showing that mLaSDI outperforms traditional LaSDI variants both in accuracy and computational efficiency.

\subsection{Outline}
In section \ref{sec:Preliminaries} we describe our problem setting and the LaSDI algorithm. In section \ref{sec:mLaSDI} we describe the mLaSDI algorithm. In section \ref{sec:examples} we provide numerical examples, and in section \ref{sec:disc} we provide our concluding remarks.

\section{LaSDI Framework}
\label{sec:Preliminaries}
Before discussing mLaSDI, we must briefly describe the LaSDI algorithm.
Most variations to the LaSDI algorithm differ only in their choice of dynamics identification in the latent space, or the choice of interpolation scheme for learned dynamics.
In either case, the mLaSDI framework we introduce in Section~\ref{sec:mLaSDI} applies broadly to all such variants.

\subsection{Governing Equations}

We consider parameterized ODEs of the form
\begin{equation}
	\frac{\id }{ \id t }\vc u (t; \pmb \mu) = \vc f (\vc u ; \pmb \mu), \quad \vc u(0; \pmb \mu) = \vc u_0 (\pmb \mu),
	\label{eq:govode}
\end{equation}
where $\vc u : \R^{+} \to \R^{N_u}$ is the state vector, $\vc f: \R^{N_u} \to \R^{N_u}$ is a (possibly nonlinear) vector-valued function, and $\pmb \mu \in \mathcal D \subset \R^{N}$ is an input parameter. 
This parameter may affect either the physics of the simulation or the initial condition. 
The goal of LaSDI is to rapidly predict the evolution of the state vector $\vc u$ for unseen input parameter values.

We assume no knowledge of the function $\vc f$, and only consider snapshots of the state vector $\vc u$ at discrete times $t_i, \ i = 0, 1, ..., N_t$.
For simplicity, we assume uniform timestep size $\Delta t = t_{i+1} - t_i$.
For a given input parameter $\pmb \mu^{(i)}$, we form snapshots of the state vector into the training data matrix $U^{(i)} = [\vc u^{(i)}(0), \vc u^{(i)}(t_1), ... , \vc u^{(i)}(t_{N_t}) ] \in \R^{(N_u \times (N_t+1))} $. 
Concatenating the snapshots from each of our $N_{\mu}$ training parameters yields the tensor
\begin{equation}
	\pmb U = [  U^{(1)},  U^{(2)}, ...,  U^{(N_{\mu})} ]\in \R^{N_{\mu} \times (N_t + 1) \times N_u},
\end{equation}
where $\pmb U$ is our training data tensor.

\subsection{Data compression}
Often the state dimension $N_u$ is large, and so we compress the data to facilitate easier analysis and modeling.
In this manuscript, we always compress the training data $\pmb U$ with an autoencoder \cite{Hinton2006}, as this generally allows for smaller latent dimension and higher reconstruction accuracy than linear methods.

These autoencoders consist of an encoder map which compresses the data, and a decoder map projects the compressed data back to the original state space. 
More precisely, the encoder learns a map $\mathcal G_{\text{enc}} : \R^{N_u} \to \R^{N_z}$ and the decoder learns the map $\mathcal G_{\text{dec}} : \R^{N_z} \to \R^{N_u}$, where $N_z \ll N_u$.
Applying the encoder to each snapshot gives the latent representation
\begin{equation}
	\vc z^{(i)}(t_k) := \mathcal{G}_{\mathrm{enc}}\bigl(\vc u^{(i)}(t_k)\bigr) \in \R^{N_z}.
\end{equation}
We can similarly map the training data matrix $U^{(i)}$ to a low dimensional compressed training to obtain a compressed training data matrix and tensor given by
\begin{align}
	Z^{(i)} &= \bigl[\vc z^{(i)}(t_0),\, \vc z^{(i)}(t_1),\, \ldots,\, \vc z^{(i)}(t_{N_t})\bigr], \\
	\pmb Z &= \bigl[Z^{(1)},\, Z^{(2)},\, \ldots,\, Z^{(N_\mu)}\bigr] \in \R^{N_\mu \times (N_t + 1) \times N_z}.
\end{align}
We also define the autoencoder reconstruction of our training data tensor $\hat{\pmb U}$, given by 
\begin{equation}
	\hat{\pmb U} := \mathcal G_{\text{dec}}\mathcal G_{\text{enc}} \pmb U.
\end{equation}
Our autoencoder then attempts to minimize the reconstruction loss 
\begin{equation}
	\mathcal L_{\text{AE}} ( \pmb \theta_\text{enc} , \pmb \theta_\text{dec} ) = \| \pmb U - \hat{\pmb U} \|^2,
\end{equation}
where  $\| \cdot \|$ is the element-wise $\ell^2$-norm.

\subsection{Latent Space Dynamics Identification}
\label{subsec:LaSDI}

Autoencoders allow us to compress the training data, but they do not provide any interpretability in the latent space.
To address this, LaSDI imposes user-specified dynamics in the latent space which the autoencoder attempts to satisfy.

Applying SINDy \cite{Brunton2016}, we approximate the time derivative of our compressed snapshots with a library of functions satisfying
\begin{equation}
	\dot{Z}^{(i)} \approx \dot{\hat{Z}}^{(i)} := \Theta ( Z^{(i)} ) \Xi^{(i)},
\end{equation}
where $\dot{\hat{Z}}^{(i)}$ is the approximated time derivative, $\Theta ( Z^{(i)} )$ is a user-chosen library of functions and $\Xi^{(i)}$ is a coefficient matrix to be determined. 
While we are free to choose any terms in our library, a common choice for LaSDI variants is to assume linear dynamics \cite{Bonneville2024,He2023} so that the library and coefficients are given by
\begin{equation}
	\Theta ( Z^{(i)} ) = (\vc 1 \ (Z^{(i)})^\top ), \quad \Xi^{(i)} = ( \vc b^{(i)} \  A^{(i)} )^\top,
 \label{eq:lineardynam}
\end{equation}
where $\vc 1 \in \R^{N_t}$ is a vector with all entries equal to 1, and we must solve for the coefficients $\vc b^{(i)} \in \R^{N_z}$ and  $A^{(i)} \in \R^{N_z\times N_z}$.
We will also assume linear dynamics for all LaSDI examples described in this paper.
This choice is due to both simplicity and stability, discussed further in \ref{app:2ndorderSINDy}.

To determine the coefficients $\Xi^{(i)}$, we solve the least squares problem
\begin{equation}
	\min_{\vc b^{(i)} \in \R^{N_z}, A^{(i)} \in \R^{N_z\times N_z} } \| \dot{ Z}^{(i)}  - (\vc 1 \ (Z^{(i)})^\top ) ( \vc b^{(i)} \  A^{(i)} )^\top  \|^2.
	\label{eq:SINDY_LSP}
\end{equation}
While many applications of SINDy enforce sparsity in the coefficient matrix (as the algorithm's name implies), we use a dense coefficient matrix.
This allows us to retain greater representation power in the latent space.
Additionally, if we enforced sparsity, the SINDy coefficients corresponding to each training parameter would yield a different sparsity pattern. 
This sparsity would then be violated with the interpolation scheme described in Section~\ref{subsec:GPinterp}.

After solving the least squares problem~\eqref{eq:SINDY_LSP} for each of our $N_\mu$ simulations, we form tensors for our SINDy approximations of the latent space dynamics and the corresponding coefficient matrices
\begin{align}
	\dot{\hat {\pmb Z}} &= [  \dot{\hat Z} ^{(1)}  \ \dot{\hat Z} ^{(2)} \ ... \ \dot{\hat Z} ^{(N_{\mu})} ]\in \R^{N_{\mu} \times (N_t + 1) \times N_z} \\
	 \pmb \Xi &= [  \Xi^{(1)} \  \Xi^{(2)} \ ... \   \Xi^{(N_{\mu})} ]\in \R^{N_{\mu} \times N_z \times (N_z + 1)}.
\end{align}
We can now define our dynamics identification loss as the difference between the true and approximated time derivative
\begin{equation}
	\mathcal L_{\text{DI}} ( \pmb \Xi) = \| \dot{ \pmb Z} - \dot{ \hat{ \pmb Z} }\|^2.
\end{equation}
{ In this manuscript we approximate the time derivative $\dot{ \pmb Z}$ using a second-order central-difference scheme here, although a weak form approach has also been studied for the case of noisy data \cite{Messenger2021, Tran2024}.}

For the final term of our loss function, we penalize the norm of our SINDy coefficients to obtain the loss function for LaSDI
\begin{equation}
	\mathcal L ( \pmb \theta_\text{enc} , \pmb \theta_\text{dec}, \pmb \Xi ) = \mathcal L_{\text{AE}} ( \pmb \theta_\text{enc} , \pmb \theta_\text{dec} )  + \beta_1 \mathcal L_{\text{DI}} ( \pmb \Xi) + \beta_2 \| \pmb \Xi \|^2
    \label{eq:lasdiloss}
\end{equation}
{ The coefficient penalization term acts as a regularizer to ensure the coefficients of our ODEs are small, which helps to prevent overfitting and empirically improves results when interpolating our ODEs to unseen input parameters.}
By training with the loss function~\eqref{eq:lasdiloss}, our autoencoder attempts to accurately reconstruct the training data while satisfying linear dynamics in the latent space. 

We note that higher-order terms can be included in the SINDy library~\eqref{eq:lineardynam}, but as demonstrated in~\ref{app:2ndorderSINDy}, such terms can cause instabilities during prediction without improving accuracy.
In practice, the nonlinearity of the autoencoder is often sufficient to model complex dynamics even when the latent space dynamics are restricted to be linear.


\subsection{Gaussian Process Interpolation}
\label{subsec:GPinterp}

After training, we require a method to obtain SINDy coefficients $\Xi^{(*)}$ for a new test parameter $\pmb \mu^{(*)}$. 
Here, we use the Gaussian Process (GP) interpolation scheme of Gaussian Process-based LaSDI (GPLaSDI).
We refer the reader to \citet{Bonneville2024} for full details of the interpolation and only provide a brief description here.

For each entry of the coefficient matrix, we fit a GP to the training data to find the mapping
\begin{equation}
    \mathcal{GP}_{\pmb \Xi}: \pmb \mu^{(*)} \mapsto \{m( \Xi^{(*)}), s( \Xi^{(*)})\},
\end{equation}
where $m$ is the predictive mean of $ \Xi^{(*)}$ and $s$ the predictive standard deviation.
When predicting for a test parameter $\mu^{(*)}$, we have two choices evolving the compressed initial condition.
Our first choice is to evolve in the latent space using the mean SINDy coefficients provided by the GPs. 
However, we can also sample the GPs for each SINDy coefficient several times to get multiple predictions for latent space dynamics.
Decoding these predictions provides different reconstructions of the state vector for each sample.
Calculating the variance of these predictions, we obtain a measure of uncertainty for our GPLaSDI predictions without any knowledge of the governing equations.

\section{mLaSDI}
\label{sec:mLaSDI}

Training with the loss function~\eqref{eq:lasdiloss} requires the autoencoder to simultaneously reconstruct the training data and produce latent trajectories satisfying linear ODEs.
These two objectives heavily restrict the representations our autoencoder can learn compared to networks trained without dynamics constraints.
As we demonstrate in Section \ref{sec:examples}, placing these restrictions on the autoencoder may inhibit us from learning accurate representations of the training data, particularly for complex or high-frequency dynamics.

\begin{figure}
  \centering
  \includegraphics[width = \linewidth]{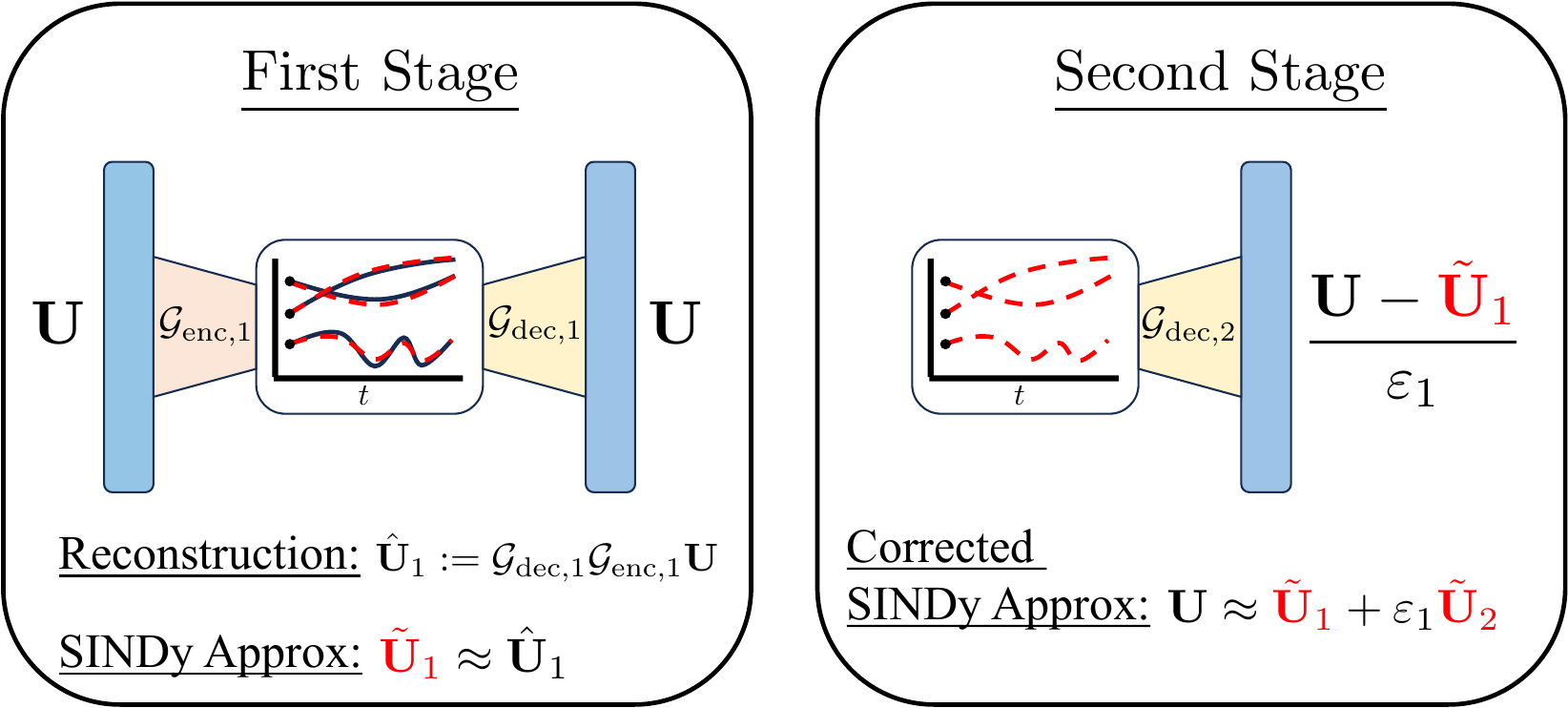}
  \caption{Schematic of mLaSDI. The first stage learns an autoencoder which is trained to reconstruct the data while learning SINDy dynamics in the latent space. The latent trajectories (solid black lines) are approximated using SINDy (dashed red lines). In the second stage, a new decoder maps the SINDy-approximated latent trajectories to the normalized residual from the first stage. The final reconstruction combines outputs from both decoders.}
  \label{fig:mlasdi_scheme}
\end{figure}

Multi-stage Latent Space Dynamics Identification (mLaSDI) addresses this limitation by training additional decoders that correct the residual error from earlier stages. 
Using multiple stages, we are able to improve prediction errors and reduce training time compared to using a single autoencoder. 
This also allows us to perform less hyperparameter searching, as we are able to implement multiple smaller models to achieve high accuracy. 
Figure~\ref{fig:mlasdi_scheme} provides a diagram detailing the mLaSDI framework.

\subsection{Multi-stage training}

The first stage of mLaSDI is identical to the LaSDI algorithm described in Section~\ref{sec:Preliminaries}.
For training data $\pmb{U}$, we obtain an autoencoder reconstruction of training data from the first stage $\hat{ \pmb{U}}_1$.
We can also obtain SINDy reconstructions of the training data by decoding our SINDy approximations of the latent space dynamics, i.e.
\begin{equation}
    \tilde{\pmb U}_1 := \mathcal{G}_{\mathrm{dec},1}(\hat{\pmb Z}),
\end{equation}
where the subscript refers to the fact that this is our first stage of training and ${\hat{\pmb{Z}}}$ is our SINDy approximation of the latent trajectories ${{\pmb{Z}}}$.

After training the first stage, it may be the case that the training data is not reconstructed accurately by the SINDy approximation $\tilde{\pmb U}_{1}$.
When this happens, we have no hope of accurately predicting solutions for unseen input parameters.
So, with the goal of improving our SINDy reconstruction accuracy, we consider the residual between the true training data ${\pmb U}$ and the SINDy reconstructions of the training data $\tilde{\pmb U}_{1}$,
\begin{equation}
    \pmb R_{1} = {\pmb U} - \tilde{\pmb U}_{1}.
    \label{eq:res1}
\end{equation}
In the mLaSDI framework, we introduce a second decoder which takes the first stage's SINDy latent space trajectories as input, and attempts to reconstruct the normalized residual~\eqref{eq:res1}. 
More precisely, we fix model parameters of the first autoencoder, and learn a mapping for the second stage $\mathcal G_{\text{dec}, 2} : \R^{N_z} \to \R^{N_u}$ by minimizing
\begin{equation}
    \mathcal L_{\mathrm{dec},2} ( \pmb \theta_{\mathrm{dec},2} ) =  \| \eps_{1}^{-1} \pmb R_{1} - \mathcal G_{\mathrm{dec},2} \hat{\mathbf{Z}} \|^2, \quad \eps_1 = \mathrm{std}(\mathbf R_1).
    \label{eq:mlasdi_loss}
\end{equation}
Here, $\eps_{1}$ is a scalar which normalizes the residual to have unit standard deviation.
{
This scaling is consistent with the approach of \citet{Wang2024}.
We treat the residual from the first stage as fixed, and only scale by standard deviation as this empirically helped gradient flow when training our models.
}
With this approach we can sequentially train multiple models. 
{ For each stage
$k \ge 2$, after training stages $1,\dots,k-1$, we define the residual
\begin{align}
\mathbf R_{k-1} := \mathbf U - \left(\tilde{\mathbf U}_1
+ \sum_{i=2}^{k-1} \epsilon_{i-1}\tilde{\mathbf U}_i\right),
\end{align}
where the $k^\mathrm{th}$ stage decoded SINDy trajectory is
\begin{align}
\tilde{\mathbf U}_k := G_{\mathrm{dec},k}(\hat{\mathbf Z};\boldsymbol\theta_{\mathrm{dec},k}).
\end{align}
We train the $k^\mathrm{th}$ decoder to minimize the normalized residual from
the previous stages,
\begin{align}
\mathcal L_{\mathrm{dec},k}(\boldsymbol\theta_{\mathrm{dec},k})
= \left\| \epsilon_{k-1}^{-1}\mathbf R_{k-1} - \tilde{\mathbf U}_k \right\|^2,
\end{align}
where the scalar $\epsilon_{k-1} = \mathrm{std}(\mathbf R_{k-1})$ normalizes
the standard deviation of the residual to 1.

After training stage $k$, we update the residual by
\begin{align}
\mathbf R_k := \mathbf U - \left(\tilde{\mathbf U}_1
+ \sum_{i=2}^{k} \epsilon_{i-1}\tilde{\mathbf U}_i\right).
\end{align}
}
The full mLaSDI approximation is then
\begin{equation}
\mathbf U \approx \tilde{\mathbf U}_1 + \epsilon_1 \tilde{\mathbf U}_2
+ \epsilon_2 \tilde{\mathbf U}_3 + \cdots + \epsilon_{n-1}\tilde{\mathbf U}_n.
\end{equation}

Crucially, all stages share the same latent trajectories $\hat{\mathbf Z}$, preserving the interpretability of the learned dynamics.

We will show in Section~\ref{sec:examples} that introducing one stage of training is enough to achieve significant increase in reconstruction and prediction accuracy { for the examples considered in this manuscript}.

{

\subsection{Error propagation for LaSDI and mLaSDI}

In this section, we first derive a continuous-time error estimate for LaSDI and then establish a stagewise monotonicity property for mLaSDI on the training data. 
The first result makes explicit how the state-space error depends
on both the autoencoder reconstruction and the mismatch in the learned latent dynamics. 
The second result shows that, once the shared latent trajectories are fixed, adding an additional mLaSDI stage cannot increase the training residual.

For simplicity of notation, we consider a single encoder $E : \R^N \to \R^{n_z}$ and decoder $D: \R^{n_z} \to \R^N$.
Let $\vc u : [0,T] \to \R^N$
denote a solution of our governing ODE \eqref{eq:govode}, which has the corresponding latent trajectory
$\vc z(t) := E(\vc u(t)) \in \R^{n_z}$.
Also let $\hat{ \vc f } : \mathbb{R}^{n_z} \to \R^{n_z}$
denote the learned latent dynamics, and let $\hat{\vc z}$ solve the ODE
\begin{equation}
  \dot{\hat{\vc z}}(t) = \hat{ \vc f }(\hat{\vc z}(t)), \quad \hat{\vc z}(0) = \vc z(0).
  \label{eq:latent_approx_ode}
\end{equation}
The corresponding LaSDI reconstruction is
\begin{equation}
  \hat{\vc u}(t) := D(\hat{\vc z}(t)).
\end{equation}

Throughout this subsection, $\|\cdot\|$ denotes the Euclidean norm in the
relevant finite-dimensional space.
We can now state our error bound rigorously in the following proposition.

\begin{prop}
  \label{prop:errorbound}
Assume that the decoder $D$ is Lipschitz on a set $\Omega \subset \R^{n_z}$ containing
$\vc z([0,T]) \cup \hat{\vc z}([0,T])$, i.e.,
\begin{equation}
    \|D(\vc z_1) - D(\vc z_2)\| \leq L_D \|\vc z_1 - \vc z_2\|, \qquad \vc z_1, \vc z_2 \in \Omega,
\end{equation}
for some constant $L_D > 0$. Define the autoencoder reconstruction error
\begin{equation}
    \eps_{\mathrm{AE}} := \sup_{t \in [0,T]} \|\vc u(t) - D(\vc z(t))\| = \sup_{t \in [0,T]} \|\vc u(t) - D(E(\vc u(t)))\|.
\end{equation}
Also define the latent dynamics model error
\begin{equation}
    \vc r(t) := \dot{\vc z}(t) - \hat{ \vc f }(\vc z(t)),
\end{equation}
and the corresponding maximum dynamics model error
\begin{equation}
    \eps_{\mathrm{Dyn}} := \sup_{t \in [0,T]} \|\vc r(t)\|.
\end{equation}
If $\hat{ \vc f }$ is Lipschitz on the same set with constant $L_f$, then for every
$t \in [0,T]$ we have
\begin{equation}
    \|\vc u(t) - \hat{\vc u}(t)\| \leq \eps_{\mathrm{AE}} + L_D \int_0^t e^{L_f(t-s)} \|\vc r(s)\| \, ds.
    \label{eq:lasdi_bound_general}
\end{equation}
Consequently,
\begin{equation}
    \|\vc u(t) - \hat{\vc u}(t)\| \leq\eps_{\mathrm{AE}} + L_D \eps_{\mathrm{Dyn}} \frac{e^{L_f t} - 1}{L_f},
    \label{eq:lasdi_bound_sup}
\end{equation}
provides a bound for the SINDy approximation error in our LaSDI models.
\end{prop}

\begin{proof}
  See \ref{app:prop1_proof}
\end{proof}

Proposition \ref{prop:errorbound} separates the LaSDI approximation error into error introduced by the autoencoder and the error introduced by the learned latent dynamics. 
Although the bound \eqref{eq:lasdi_bound_sup} is not necessarily tight, it shows that both effects contribute to the final reconstruction error, and that improving either term improves the bound. 
This interpretation is consistent with the structure of the loss \eqref{eq:lasdiloss}, which penalizes both reconstruction error and latent-dynamics mismatch.

We state Proposition \ref{prop:errorbound} in continuous time because the argument is cleaner, but the same interpretation carries over to the discrete-time setting used in practice. 
An analogous result can be formulated by working with the one-step map $\Phi: \R^{n_z} \to \R^{n_z}$ associated with \eqref{eq:latent_approx_ode}, which satisfies $\hat{\vc z}_{j+1}  = \Phi(\hat{\vc z}_j) $.

We note that the Lipschitz assumptions in Proposition \ref{prop:errorbound} are mild for the decoders considered here, as feedforward neural networks with standard activations such as ReLU, Softplus, and Tanh are compositions of Lipschitz functions and thus Lipschitz on bounded sets.
The Lipschitz assumption on the latent dynamics can be violated depending on the choice of latent dynamics.
When the learned latent dynamics are linear, $\hat{\vc f}(\vc z) = A\vc z + \vc b$, we can always take $L_f = \|A\|$. 
By contrast, higher-order polynomial latent dynamics may exhibit finite-time blowup, so establishing a uniform bound over a prescribed time interval becomes more delicate.
See \ref{app:2ndorderSINDy} more discussion on why higher-order dynamics may perform worse than linear latent dynamics.

Next, we examine the stagewise behavior of mLaSDI on the training data. 
More precisely, once the first-stage SINDy latent trajectories $\hat{\vc Z}$ are fixed, each additional decoder is trained to approximate the normalized residual left by the previous stages. 
Under the idealized assumption that each stagewise optimization problem is solved globally, the training residual norm
cannot increase from one stage to the next.
Notably, we do not place any restrictions on the governing equations which generate our data or latent dynamics model.

\begin{prop}[Discrete stagewise residual monotonicity]
  \label{prop:stage_mono}
Let $\vc U$ denote the training data and let $\hat{\vc Z}$ be the fixed
SINDy-approximated latent trajectories obtained after the first stage.
Define the first-stage decoded SINDy reconstruction by
\begin{equation}
  \tilde{\vc U}_1 := G_{\mathrm{dec},1}(\hat{\vc Z}), \qquad \vc R_1 := \vc U - \tilde{ \vc U }_1.
\end{equation}

For each stage $k \ge 2$, let $\eps_{k-1} > 0$ be the residual
normalization factor (in mLaSDI, $\eps_{k-1} = \mathrm{std}(R_{k-1})$),
and define
\begin{align}
\tilde{\vc U}_k &:= G_{\mathrm{dec},k}(\hat{\vc Z};\theta_{\mathrm{dec},k}), \\ 
\tilde{\vc U}^{(k)} &:= \tilde{\vc U}_1 + \sum_{i=2}^{k} \eps_{i-1}\tilde{\vc U}_i, \qquad \vc R_k := \vc U - \tilde{\vc U}^{(k)}.
\end{align}
Assume that, for each $k \ge 2$,
$\pmb \theta_{\mathrm{dec},k}^\star$ is a global minimizer of
\begin{equation}
\mathcal L_{\mathrm{dec},k}(\pmb \theta_{\mathrm{dec},k}) := \left\| \eps_{k-1}^{-1} \vc R_{k-1} - G_{\mathrm{dec},k}(\hat{\vc Z}; \pmb \theta_{\mathrm{dec},k}) \right\|^2.
\end{equation}
Then the training residual norm is nonincreasing across stages:
\begin{equation}
  \|\vc R_k\| \le \|\vc R_{k-1}\|, \qquad k \ge 2.
\end{equation}
Equivalently,
\begin{equation}
\left\| \vc U - \tilde{\vc U}_1 - \sum_{i=2}^{k} \eps_{i-1}\tilde{\vc U}_i \right\| \le \left\| \vc U - \tilde{\vc U}_1 - \sum_{i=2}^{k-1} \eps_{i-1}\tilde{\vc U}_i \right\|.
\end{equation}

Moreover, the inequality is strict whenever the zero correction is not a minimizer of $\mathcal L_{\mathrm{dec},k}$.
\end{prop}

\begin{proof}
  See \ref{app:prop2_proof}.
\end{proof}

Proposition \ref{prop:stage_mono} is a statement about the training residual evaluated on the fixed shared latent trajectories $\hat{\vc Z}$. 
It does not, by itself, guarantee improved prediction error at unseen parameter values, since generalization still depends on how the stagewise residual corrections interpolate across parameter space. 
We also note that the assumption of a global minimizer at each stage is idealized as neural network training via stochastic gradient descent provides no guarantee of finding a global minimum.
In practice, Proposition \ref{prop:stage_mono} should be interpreted as a best-case bound on training residual reduction, rather than a guarantee. 
Empirically, we observe that the second-stage decoder consistently reduces the training residual in the experiments of Section \ref{sec:examples}, which is not surprising given that we only need to find a local minimum which is better than the zero map for mLaSDI to increase reconstruction accuracy on the training data.

}

\subsection{Architecture of the later stages}

After the first stage of training, we are free to choose any architecture for the decoders at later stages.
Our goal is to require minimal hyperparameter tuning, and so we adopt the approach of \citet{Wang2024}.
All stages have identical decoder architectures, varying only the activation functions.
For all training stages after the first, we use sine activation on the first layer.
Sinusoidal activations enable neural networks to represent high-frequency content more effectively \cite{sitzmann2019siren, Wang2024}, which is important because the residuals often contain fine-scale features that the dynamics-constrained first stage cannot capture.
Subsequent layers use the hyperbolic tangent activation.

A key element of the multistage networks in \citet{Wang2024} is the introduction of a scaling factor $\kappa$ which is used to initialize weights at later layers.
Their work focuses on multi-layer perceptrons and PINNs, where the networks learn an explicit coordinate-to-output mapping.
The scaling factor is then chosen based on the frequency of the residual, and this factor helps to accelerate learning when the residual is small.

With LaSDI, we are concerned with autoencoders that learn explicit data-to-data mappings, and so the same theory does not directly transfer.
Additionally, because we are compressing the data and { simultaneously} learning { smooth} latent dynamics, achieving the machine-precision accuracy sought after in the original work {of \cite{Wang2024}} is not feasible. 
Empirically, we find that $\kappa = 1$ provides a balance between accuracy and training efficiency.
Based on the numerical study in \ref{app:kappatesting}, we set $\kappa=1$ in all the numerical results below.
{ As we will show in Section \ref{sec:examples}, setting $\kappa=1$, and essentially omitting the scaling factor, still allows mLaSDI to achieve higher accuracy in less training time than GPLaSDI models.}




\section{Numerical examples}
\label{sec:examples}


Here, we provide three numerical examples to demonstrate the performance of mLaSDI compared to GPLaSDI. 
We will show that mLaSDI allows us to train more accurate models in less time, and is less sensitive to choice of model architecture.
Moreover, mLaSDI often outperforms larger GPLaSDI models and achieves greater parameter efficiency.
Although we focus on GPLaSDI, the mLaSDI framework applies to any LaSDI variant, and we expect similar improvements for other formulations.

All first-stage autoencoders use Softplus activation function and are trained for a fixed number of iterations using Pytorch's implementation of the ADAM optimizer \cite{Kingma2014}.
When optimizing the GP parameters for our SINDy coefficients, we use the \texttt{GaussianProcessRegressor} from scikit-learn \cite{sklearn}.
We use the Mat\'{e}rn kernel
\begin{equation}
    k(\vc x, \vc y) = \frac{1}{\Gamma(\nu)2^{\nu-1}} \left(\frac{\sqrt{2\nu}}{l} \|\vc x - \vc y \|\right)^{\nu} K_{\nu}\left(\frac{\sqrt{2\nu}}{l} \| \vc x - \vc y \|^2\right),
\end{equation}
where $\Gamma (\cdot)$ is the Gamma function, $K_\nu$ is a modified Bessel function, and we set $\nu = 1.5$.
We set the hyperparameters of the loss function~\eqref{eq:lasdiloss} to $\beta_1 = 10^{-1}, \ \beta_2 = 10^{-3}$.

We describe network architectures using the notation $N_{\mathrm{in}}$-$h_1$-$h_2$-$\cdots$-$N_z$, where $N_{\mathrm{in}}$ is the input dimension, $h_i$ are hidden layer widths, and $N_z$ is the latent dimension.
For example, the notation 1000-100-10-5 represents a fully connected network with input dimension 1000, two hidden layers (sizes 100 and 10), and output dimension 5.
The decoder always has the reverse architecture of the encoder, e.g. 5-10-100-1000.
In each example, we obtain training data reconstructions and test data predictions for input parameters by evolving latent dynamics using ODEs which correspond to the mean SINDy coefficients from our GPs.

\subsection{Multiscale oscillating system}
\label{subsec:Multiscal}


\begin{table}[t]
  \centering
  \footnotesize
  \setlength{\tabcolsep}{4pt}
  \begin{threeparttable}
    \caption{Results applying GPLaSDI and mLaSDI for toy problem~\eqref{eq:multistage_func}
      and wake shedding~\eqref{eq:NS}. mLaSDI used two training stages, and the reported
      training epochs are for both stages combined. Relative errors are the range over the
      entire time evolution for training and prediction cases. Larger GPLaSDI models are given the same number of training epochs as mLaSDI with two stages to match training epochs and compare training times.}
    \label{tab:combined}
    \begin{tabular}{@{}llccccc@{}}
      \toprule
      \textbf{Problem} &
      \textbf{Model} &
      \textbf{Arch.} &
      \textbf{Epochs} &
      \shortstack{\textbf{Model}\\\textbf{Params.}} &
      \shortstack{\textbf{Train}\\\textbf{Time (s)}} &
      \shortstack{\textbf{Rel.}\\\textbf{Error}} \\
      \midrule
      Toy Problem   & GPLaSDI         & $\dagger$Shallow    & 50,000 & 292,310 & 387.87 & 4--20\% \\
      Toy Problem   & GPLaSDI         & $\dagger$Medium     & 20,000 & 454,250 & 220.83 & 1--11\% \\
      Toy Problem   & GPLaSDI         & $\dagger$Deep       & 20,000 & 694,550 & 246.15 & 5--17\% \\
      Toy Problem   & \textbf{mLaSDI} & $\dagger$Shallow    & 20,000 & 438,760 & 120.94 & \textbf{0.1--3}\% \\
      \addlinespace
      Wake Shedding & GPLaSDI         & $\ddagger$Shallow   & 10,000 & 296,939 &  48.29 & 3--5\% \\
      Wake Shedding & GPLaSDI         & $\ddagger$Medium    &  5,000 & 453,589 &  33.31 & 4--5\% \\
      Wake Shedding & GPLaSDI         & $\ddagger$Deep      &  5,000 & 602,839 &  35.35 & 4--5\% \\
      Wake Shedding & \textbf{mLaSDI} & $\ddagger$Shallow   &  5,000 & 446,873 &  22.79 & \textbf{1--2}\% \\
      \bottomrule
    \end{tabular}
    \begin{tablenotes}[flushleft]
      \small                        
      \item \textit{Architecture notation: input -- hidden$_1$ -- hidden$_2$ -- $\cdots$ -- latent.}
      \item[$\dagger$] \textit{Toy problem architectures:}
        \textbf{Shallow} (600-200-20-10);\quad
        \textbf{Medium} (600-200-200-100-50-20-10);
        \textbf{Deep} (600-400-200-100-50-20-10).
      \item[$\ddagger$] \textit{Wake shedding architectures:}
        \textbf{Shallow} (2934-50-5);\quad
        \textbf{Medium} (2934-75-50-25-5);\quad \textbf{Deep} (2934-100-50-25-5).
    \end{tablenotes}
  \end{threeparttable}
\end{table}

\begin{figure}
  \centering
  \includegraphics[width = .49\linewidth]{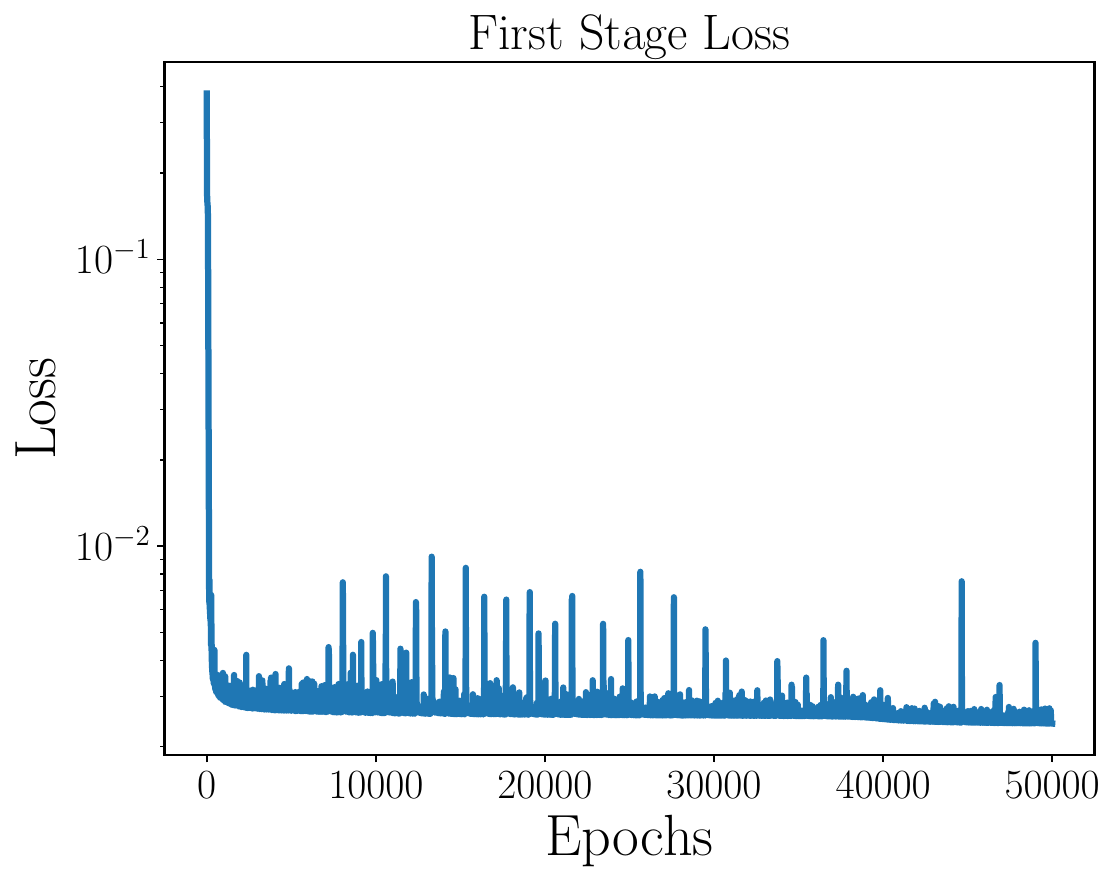}
  \includegraphics[width = .465\linewidth]{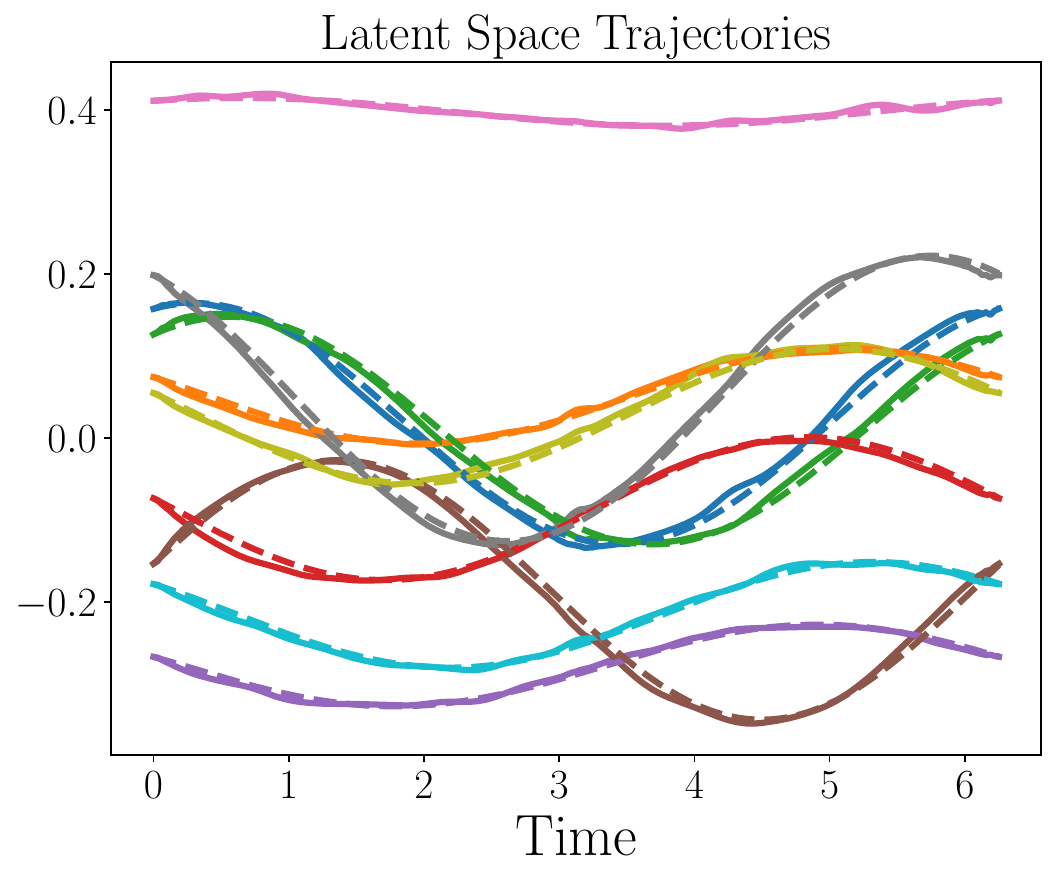}
  \caption{Applying GPLaSDI to toy problem~\eqref{eq:multistage_func}. (Left) Loss for GPLaSDI and (Right) Latent space trajectories { of our trained GPLaSDI model for training case $A = 1.4$. We plot both the autoencoder compression of the data (solid lines) and the SINDy approximation of the autoencoder trajectories (dashed lines) from compressing the initial condition and evolving using our learned ODEs.} }
  \label{fig:losses_LS}
\end{figure}

\begin{figure}
  \centering
  \includegraphics[width = \linewidth]{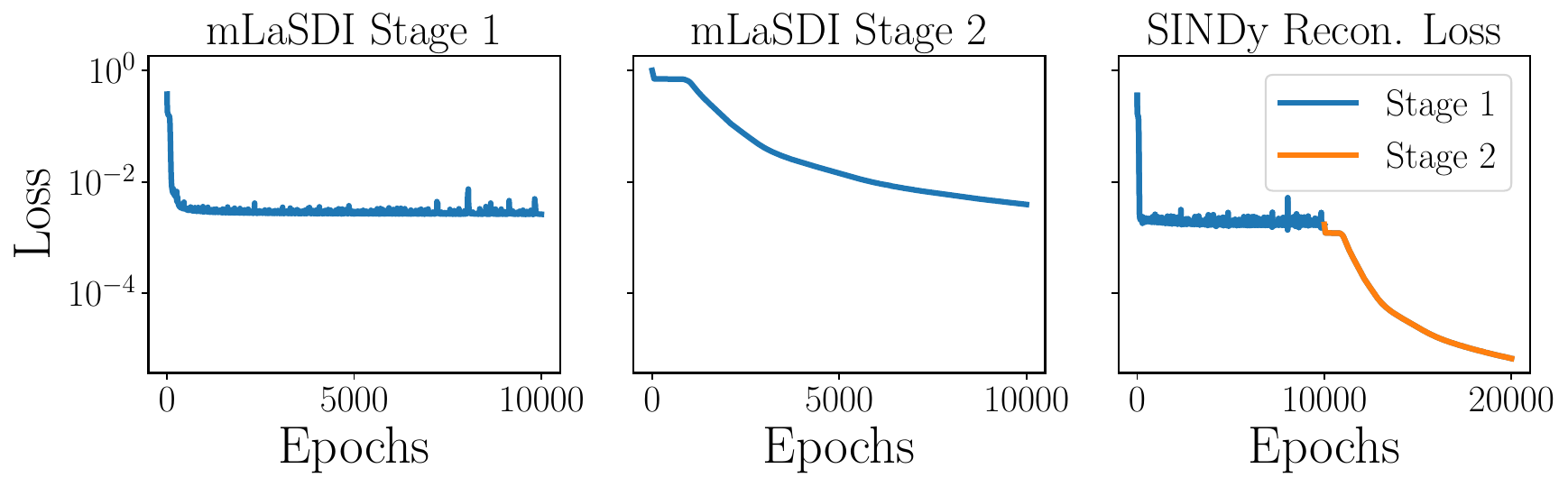}
  \caption{{ Applying mLaSDI to toy problem~\eqref{eq:multistage_func}. Introducing a second stage of training allows us to quickly decrease the error of our SINDy reconstruction for the training data after the first stage has stalled.}}
  \label{fig:mlasdi_losses}
\end{figure}

\begin{figure}
  \centering
    \includegraphics[width = \linewidth]{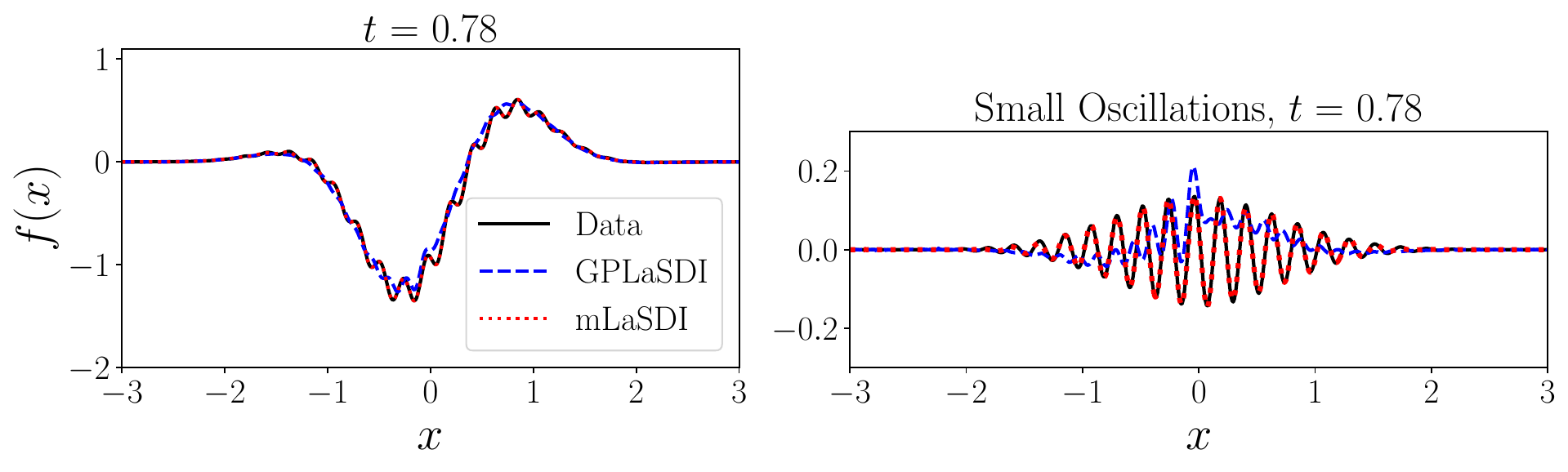}
  \caption{Applying GPLaSDI and mLaSDI with 2 stages to~\eqref{eq:multistage_func} with $A = 1.4$. (Left) Reconstruction of the training data for both methods. (Right) Results after subtracting the sine wave from the approximations to visualize how well each method approximates the small, high-frequency cosine wave. Note that the data and mLaSDI are nearly overlapping. }
  \label{fig:multiscale_data}
\end{figure}

We begin with a simple example to illustrate the spectral bias limitation of standard LaSDI. 
Consider the multiscale function
\begin{equation}
	f(x, t; A) = A \left[ \sin(2x - t) + 0.1 \cos \big( (40x + 2t) \sin(t) \big) \right] e^{-x^2},
	\label{eq:multistage_func}
\end{equation}
where the input parameter $A$ controls the amplitude of the function.
This function combines a low-frequency sine wave with a small, high-frequency cosine component to emphasize the setting where dynamics-constrained autoencoders struggle due to spectral bias.

We generate synthetic training data from the input parameter values $A \in \{1.0, 1.4\}$, and try to predict for the case $A = 1.2$.
To generate data for this toy problem, we sample~\eqref{eq:multistage_func} at 201 time steps evenly distributed on the interval $[0, 2\pi]$, and 600 spatial steps evenly distributed in the interval $[-3, 3]$.
This yields 201 snapshots of the state vectors $\vc u (t; A) \in \R^{600}$.

We train two stages of mLaSDI for 10,000 iterations each. 
For comparison, we also train one stage of GPLaSDI for 50,000 iterations to emphasize that one stage of training is insufficient to accurately capture dynamics of the data.
Our autoencoders for both mLaSDI and GPLaSDI have the same architecture, consisting of 2 hidden layers with architecture 600-200-20-10.
We then train our models for this toy problem on CPU using a 2021 MacBook Pro with Apple M1 Max Chip and 32 GB of memory.

In Figure \ref{fig:losses_LS} we plot the loss curve for GPLaSDI along with latent space trajectories for the training case $A = 1.4$. 
It is clear that training loss quickly stagnates, and the GPLaSDI model is no longer improving despite additional training. 
{ For comparison, we also plot the loss-curves for each stage of our mLaSDI model in Figure \ref{fig:mlasdi_losses}. 
Immediately, we see that the loss from tbe second stage of mLaSDI quickly decreases after the first stage has stalled.
Additionally, the SINDy reconstruction error of the training data begins to rapidly decrease while the first stage GPLaSDI model leads to no significant improvement.}

Figure~\ref{fig:multiscale_data} compares reconstructions from both methods.
From this snapshot we see that GPLaSDI and mLaSDI both capture the behavior of the low-frequency sine wave (left panel).
However, the right panel shows that, after removing the low-frequency sine component, GPLaSDI fails to accurately capture the high-frequency cosine component.
{
This failure is not due to lack of training, but a limitation of our autoencoder with smooth latent dynamics.}
In contrast, mLaSDI provides a significant improvement over GPLaSDI and accurately approximates both the large sine wave and small cosine wave.
The second-stage decoder quickly learns the high-frequency residual that the dynamics-constrained first stage struggles to represent.

\begin{figure}
  \centering
  \includegraphics[width = .9\linewidth]{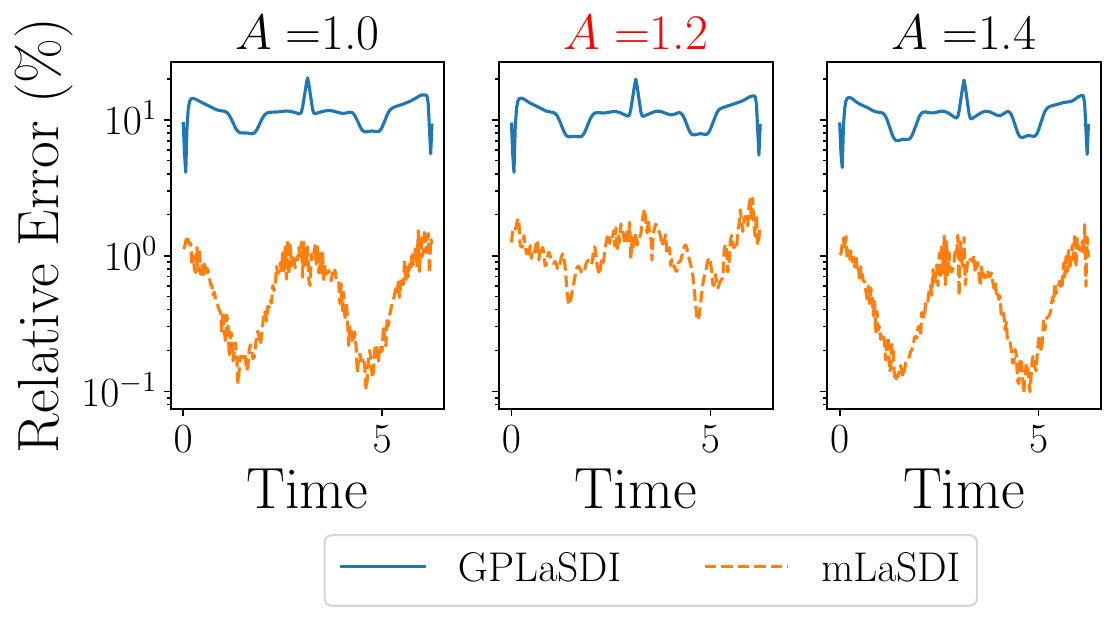}
  \caption{Relative error for GPLaSDI and mLaSDI applied to toy problem~\eqref{eq:multistage_func}. GPLaSDI is trained for 50,000 iterations while mLaSDI has 2 stages each trained for 10,000 iterations. Both methods are trained on the data from cases $A \in \{1.0, 1.4\}$, and we predict for  $A = 1.2$. }
  \label{fig:losses_sine}
\end{figure}

As we see in Figure \ref{fig:losses_sine}, GPLaSDI's failure to capture the full behavior of the function~\eqref{eq:multistage_func} leads to training and predictive relative errors of approximately 4--20\%.
mLaSDI reduces this error by an order of magnitude, achieving relative errors between 0.1--3\%.
This simple example shows how mLaSDI significantly improves the representation power of our models over base GPLaSDI.


{ We also note that mLaSDI outperforms GPLaSDI even when using a larger GPLaSDI models.
In Table \ref{tab:combined} we report results from training a GPLaSDI with a comparable number of learnable to our 2-stage model mLaSDI model, and another GPLaSDI model with significantly more learnable parameters.
For a fixed number of training epochs across these three tests, mLaSDI achieves the lowest relative errors in the least training time, emphasizing how this multi-stage approach allows us to achieve more efficient training with greater accuracy.
}

\begin{figure}
  \centering
  \includegraphics[width = .465\linewidth]{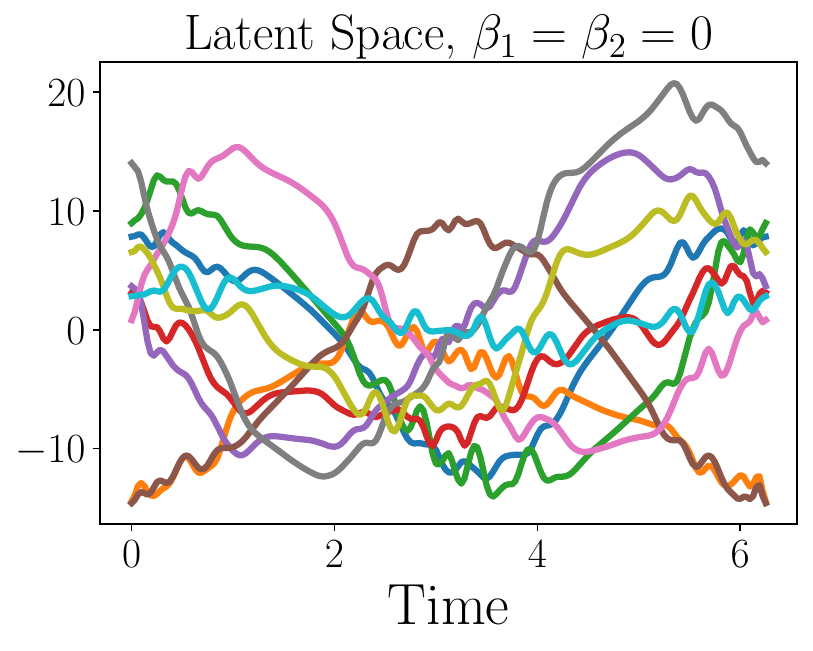}
  \includegraphics[width = .49\linewidth]{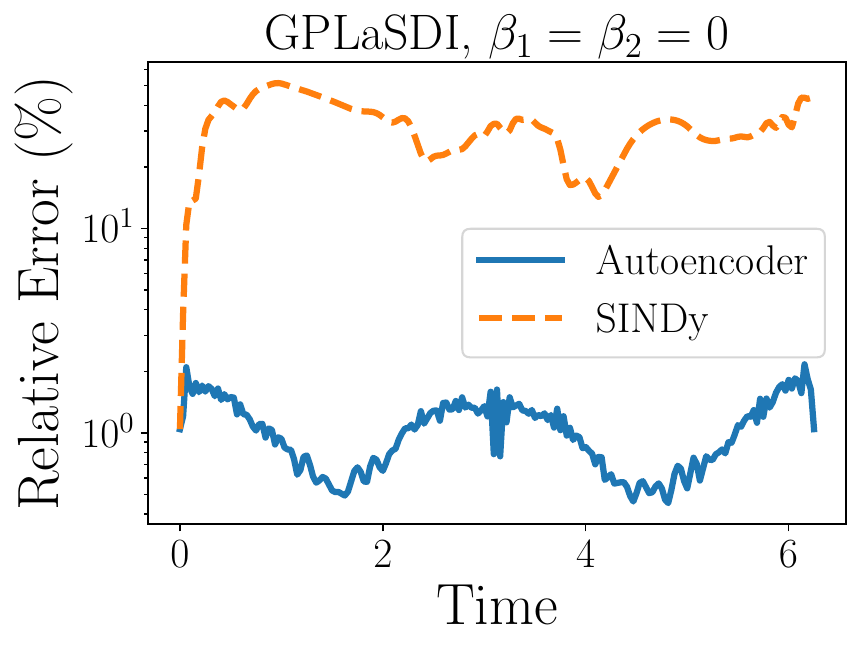}
  \caption{{ Training an autoencoder without latent dynamics penalization for toy problem~\eqref{eq:multistage_func}, i.e. setting $\beta_1 = \beta_2 = 0$ in~\eqref{eq:lasdiloss}. 
  (Left) Autoencoder latent trajectories and (Right) relative error for training case $A = 1.4$.
  While our autoencoder reconstructs the data within 2\% relative error, the latent trajectories are highly oscillatory and so the SINDy reconstruction has large relative errors (labeled GPLaSDI in plot).}}
  \label{fig:AE_nodyn}
\end{figure}

{
We conclude this example by noting that the autoencoder struggles specifically because we impose smooth latent dynamics during training.
In Figure \ref{fig:AE_nodyn} we plot the latent space trajectories and relative error when training an autoencoder with architecture 600-200-20-10 on $A \in {1, 1.4}$ without any latent dynamics (i.e. $\beta_1 = \beta_2 = 0$).
Without enforcing smooth ODEs, the autoencoder is able to learn highly oscillatory latent space trajectories, and thus achieves a relative error below 2\% on the training data.
However, relative error for the SINDy reconstruction of the training data is nearly 40\%, which is twice what we observed when training with latent dynamics in Figure \ref{fig:losses_sine}.
We emphasize that training only an autoencoder does not allow us to make predictions for unseen input parameters, and can only reconstruct the training data. 
Without the SINDy dynamics identification step, there is no mechanism to interpolate or extrapolate latent trajectories to new parameter values, and we cannot evolve the new initial condition.
}

\subsection{Unsteady Wake Flow}
\label{subsec:2dfocs}

Next we consider unsteady wake flow, governed by the incompressible Navier--Stokes equations
\begin{align}
    \dfrac{\partial \vc u}{\partial t} + (\vc u \cdot \nabla) \vc u &= - \nabla p + \nu \nabla^2 \vc u \quad \mathrm{in} \ \Omega \\
    \nabla \cdot \vc u &= 0 \quad \quad \quad \quad \quad \quad \quad \ \mathrm{in} \ \Omega,
    \label{eq:NS}
\end{align}
where $\vc u (x, y, t) = (u_1(x, y, t), u_2(x, y, t))$ is the fluid velocity, $p$ is the pressure, $\nu$ is the kinematic viscosity, and $\Omega$ is the spatial domain.
The domain is a 2D rectangle $(x, y) \in [0, 1] \times [0, 2]$, with a cylinder of radius 0.05 centered at $(x, y) = (0.5, 0.5)$.
We impose Dirichlet boundary condition at the inlet $\vc u(x, 0, t) = (0, 1)$.
We apply homogeneous Dirichlet boundary conditions on the cylinder, and natural boundary conditions on the remaining boundaries.

\begin{figure}[]
  \centering
  \includegraphics[width = 0.5\linewidth]{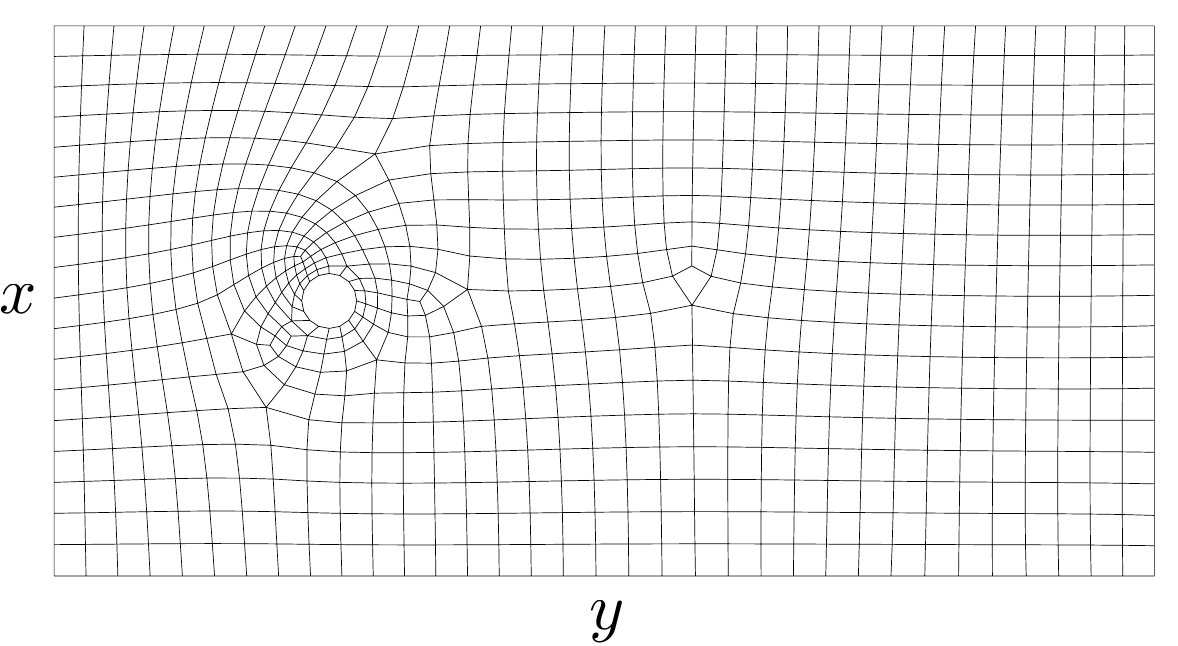}
  \caption{Mesh used to solve unsteady wake flow, rotated 90 degrees clockwise.}
  \label{fig:2dfocs_mesh}
\end{figure}

In Figure \ref{fig:2dfocs_mesh} we display our mesh, which was generated using GMSH \cite{GMSH} and has 2,934 spatial nodes.
We solve the incompressible Navier--Stokes equations using PyMFEM's Navier solver \cite{Anderson2021} with first-order elements and timestep $\Delta t = 0.001$ on a once-refined mesh (11{,}492 nodes), then downsample the solution back to the original mesh.
Each simulation is run for 10 time units to establish vortex shedding, followed by 1 additional time unit from which we extract 101 snapshots (every 10th timestep).
This yields state vectors $\vc u (t; \nu) \in \R^{2934}$ for $\nu \in \{0.00032, 0.00043, 0.00054\}$.
We train mLaSDI and GPLaSDI on the velocity magnitude of snapshots generated for $\nu \in \{0.00032, 0.00054\}$ and predict for $\nu = 0.00043$

Autoencoders for unsteady wake flow were trained on CPU using a 2021 MacBook Pro with Apple M1 Max Chip and 32 GB of memory. 
Table~\ref{tab:combined} contains architectures for the training of GPLaSDI and mLaSDI.
Similar to the previous example, we train two stages of mLaSDI for 2,500 iterations each. 
For comparison, we also train one stage of GPLaSDI for 10,000 iterations.
Our autoencoders for both methods have the same architecture, consisting of 1 hidden layer with architecture 2934-50-5.

\begin{figure}
  \centering
  \includegraphics[width = .9\linewidth]{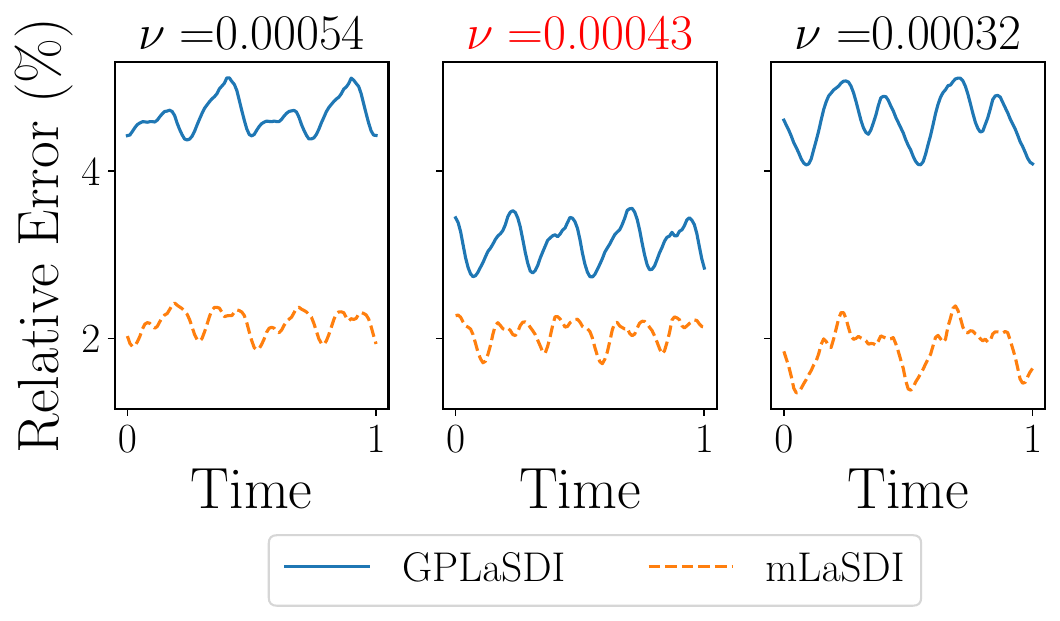}
  \caption{Relative error for GPLaSDI and mLaSDI applied to unsteady wake flow. GPLaSDI is trained for 10,000 iterations while mLaSDI has 2 stages each trained for 2,500 iterations. Both methods are trained on the data from cases $\nu \in \{0.00032, 0.00054\}$, and we predict for $\nu = 0.00043$.}
  \label{fig:2dfocs_err}
\end{figure}

\begin{figure}
  \centering
  \includegraphics[width = .9\linewidth]{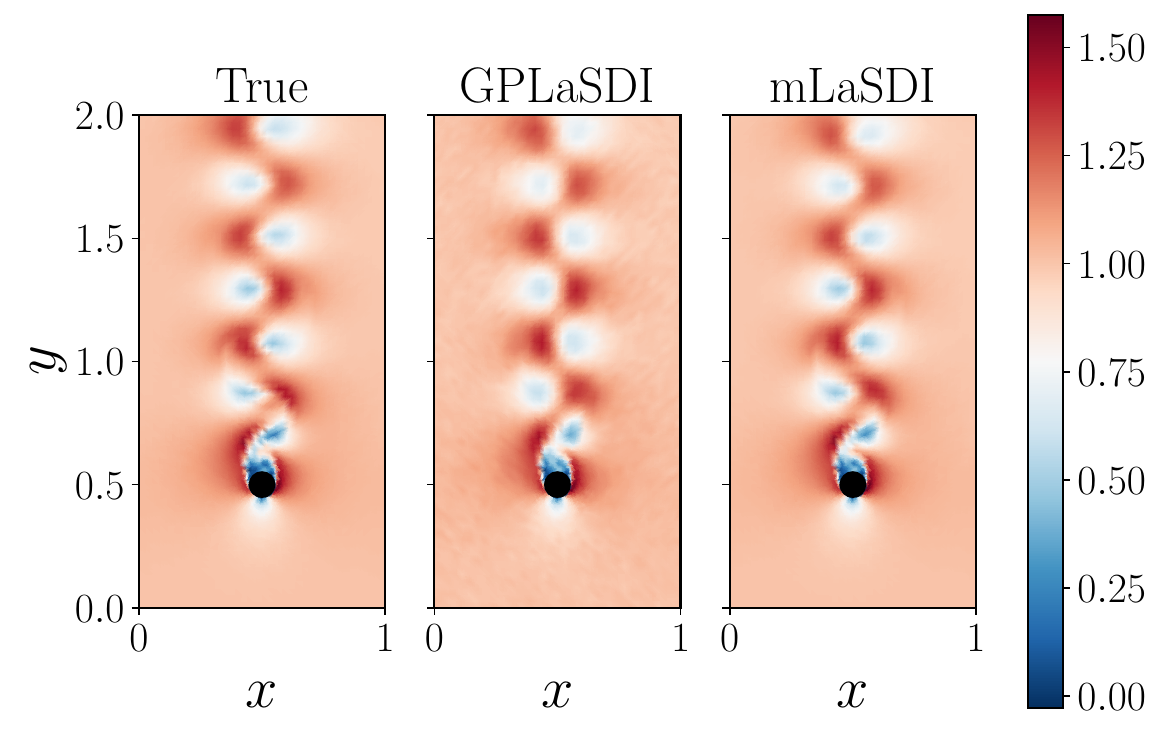}
  \caption{Snapshot of velocity magnitude for wake shedding~\eqref{eq:NS} problem at $t=0.5$ for the interpolation case $\nu = 0.00043$.}
  \label{fig:2dfocs_comp}
\end{figure}

Figure \ref{fig:2dfocs_err} shows relative errors of GPLaSDI and mLaSDI for our three kinematic viscosities.
In this case, both methods produce reasonable relative errors within 5\%, with mLaSDI reducing errors by only 1--3\% compared to GPLaSDI.
While this improvement is more modest than the order-of-magnitude gains observed for the multiscale problem, the qualitative differences are still notable.
In Figure \ref{fig:2dfocs_comp} we plot a snapshot of the solutions to demonstrate that there is still a visible qualitative difference between GPLaSDI and mLaSDI.
The vortices of the GPLaSDI prediction are more circular than the true vortices, and additionally the background velocity field has some visible noise in the error.
In contrast, mLaSDI produces more elongated vortices which are closer in shape to the true data, while also producing a smoother background velocity field.
So, mLaSDI still provides a significant qualitative improvement over GPLaSDI for this problem even though the quantitative decrease in relative error is only 1--3 percentage points. 


%

{ As in the toy example, we again compare our mLaSDI model to a similar size and larger GPLaSDI model.
As summarized Table \ref{tab:combined}, mLaSDI is consistently able to achieve the highest accuracy with less training time than one-stage GPLaSDI models.
}

\subsection{Two stream plasma instability}
\label{sec:Vlasov}

\begin{table}[t]
  \caption{Model architectures used to train GPLaSDI and mLaSDI for 1D-1V Vlasov experiments.}
  \label{tab:1d1v}
  \centering
  \footnotesize
  \begin{tabular}{lll}
    \toprule
    \textbf{Component} & \textbf{Choices} & \textbf{Description} \\
    \midrule
    mLaSDI Hidden Layers & 50, 500, 1000, 500-50, 1000-500-50 & Fully connected layers \\ 
    GPLaSDI Hidden Layers & 75, 750, 1500, 750-75, 1500-750-75 & Fully connected layers \\ 
    Latent Dimension & 4, 5, 6, 7 & Bottleneck dimension \\ 
    GPLaSDI Training Config & 25k, 50k, 75k, 100k & Training checkpoints \\
    mLaSDI Training Config. & (25k,25k), (25k,50k), (50k,50k), (75k,25k) & (Stage 1, Stage 2) iter. pairs \\
    \bottomrule
  \end{tabular}
\end{table}

Our previous examples demonstrated that mLaSDI achieves improved accuracy over GPLaSDI models with identical architectures.
Here, we show that mLaSDI with smaller networks can outperform larger GPLaSDI models, achieving higher accuracy in less training time.

For our final example, we consider the 1D-1V Vlasov equation
\begin{equation}
	\begin{cases}
		\displaystyle \frac{\partial f}{\partial t} +  \frac{\partial}{\partial x}(v f) + \frac{\partial }{\partial v} \bigg( \frac{\id \Phi }{\id x} f \bigg) = 0, \quad t \in (0, 5], \ x \in [0, 2\pi], \ v \in [-7, 7] \\
		\displaystyle \frac{\id^2 \Phi }{\id x^2} = \int_v f \ \id v, \\
        f(x, v, 0; \pmb \mu) =  \displaystyle \frac{8}{\sqrt{2 \pi T}} \bigg[ 1 + 0.1\cos(kx) \bigg] \bigg[ \exp \left( - \frac{(v - 2)^2}{2T} \right) +  \exp \left( - \frac{(v + 2)^2}{2T} \right) \bigg].
	\end{cases}
	\label{eq:Vlasov}
\end{equation}
Here, $f(x, v, t)$ is the electron distribution function which depends on space $x$ and velocity $v$. 
The function $\Phi(x)$ is the electrostatic potential. 
Our initial condition consists of two streams centered at $v = \pm 2$, and is parameterized by $\pmb \mu = (T, k)^\top$.
These parameters control the width and periodicity of the two initial streams.
This configuration exhibits the two-stream instability, causing the beams to merge and form vortical structures in phase space.

We solve the 1D-1V Vlasov equation using HyPar \cite{Hypar} with a WENO spatial discretization scheme \cite{Jiang1996} and fourth-order Runge--Kutta time integration scheme with timestep $\Delta t = 0.005$. 
We consider the parameter domain $(T, k) \in [0.9, 1.1] \times [1.0, 1.2]$, discretized with spacing $\Delta T = \Delta k = 0.01$ to produce 441 parameter combinations.
To generate data, we sample the solution at every timestep from a uniform $64 \times 64$ grid in the space-velocity field to obtain 251 snapshots of our state vectors $\vc u (t ; \pmb \mu) \in \R^{4096}$.
We then take snapshots from 25 full-order model solutions on a $5\times5$ uniform grid as training data, and attempt to predict for the other parameters.

Training is performed on an NVIDIA V100 GPU via the LLNL Lassen cluster.
We train mLaSDI using a wide range of architectures, latent space dimensions, and training iterations, with all configurations given in Table \ref{tab:1d1v}.
Hidden layers of GPLaSDI models are widened by a factor of 1.5 relative to mLaSDI.
The second stage of mLaSDI requires training a new decoder, so a 2-stage model has approximately $1.5\times$ the number of parameters as a single-stage model.
For this experiment, the hidden layers for GPLaSDI were widened by a factor of $1.5$ compared to mLaSDI, leading to GPLaSDI models which have approximately $ 2.25 \times$ more parameters than the first stage of mLaSDI and $ 1.5 \times$ the parameters as our 2-stage mLaSDI models.
Our goal here is to demonstrate that the multi-stage training achieves superior performance with fewer model parameters.

To evaluate our results, we define the relative error between a full-order simulation $\vc u(t ; \pmb \mu^{(*)})$ and the mLaSDI approximation $\tilde{ \vc u}(t ; \pmb \mu^{(*)} )$ as 
\begin{equation}
    r^{(*)} := \max_{\substack{j = 0,...,N_t}} \frac{ \| \vc u(t_j; \pmb \mu^{(*)}) - \tilde{ \vc u}(t_j;\pmb \mu^{(*)}) \| }{ \| \vc u(t_j ; \pmb \mu^{(*)}) \| },
    \label{eq:maxerrs}
\end{equation}
where $\| \cdot \|$ is the usual Euclidean norm.

\begin{figure}
  \centering
      \includegraphics[width=.49\linewidth]{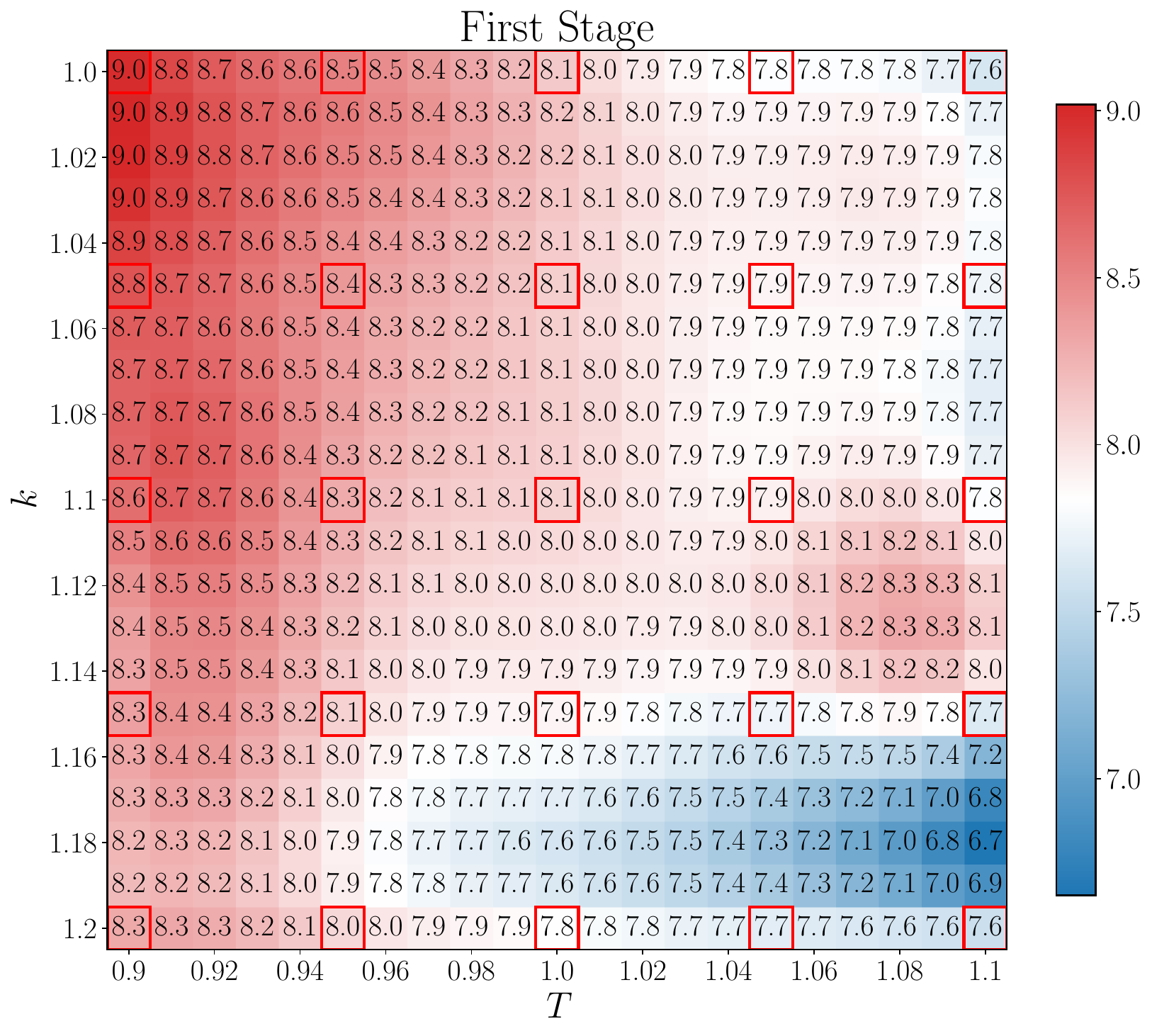}
      \includegraphics[width=.49\linewidth]{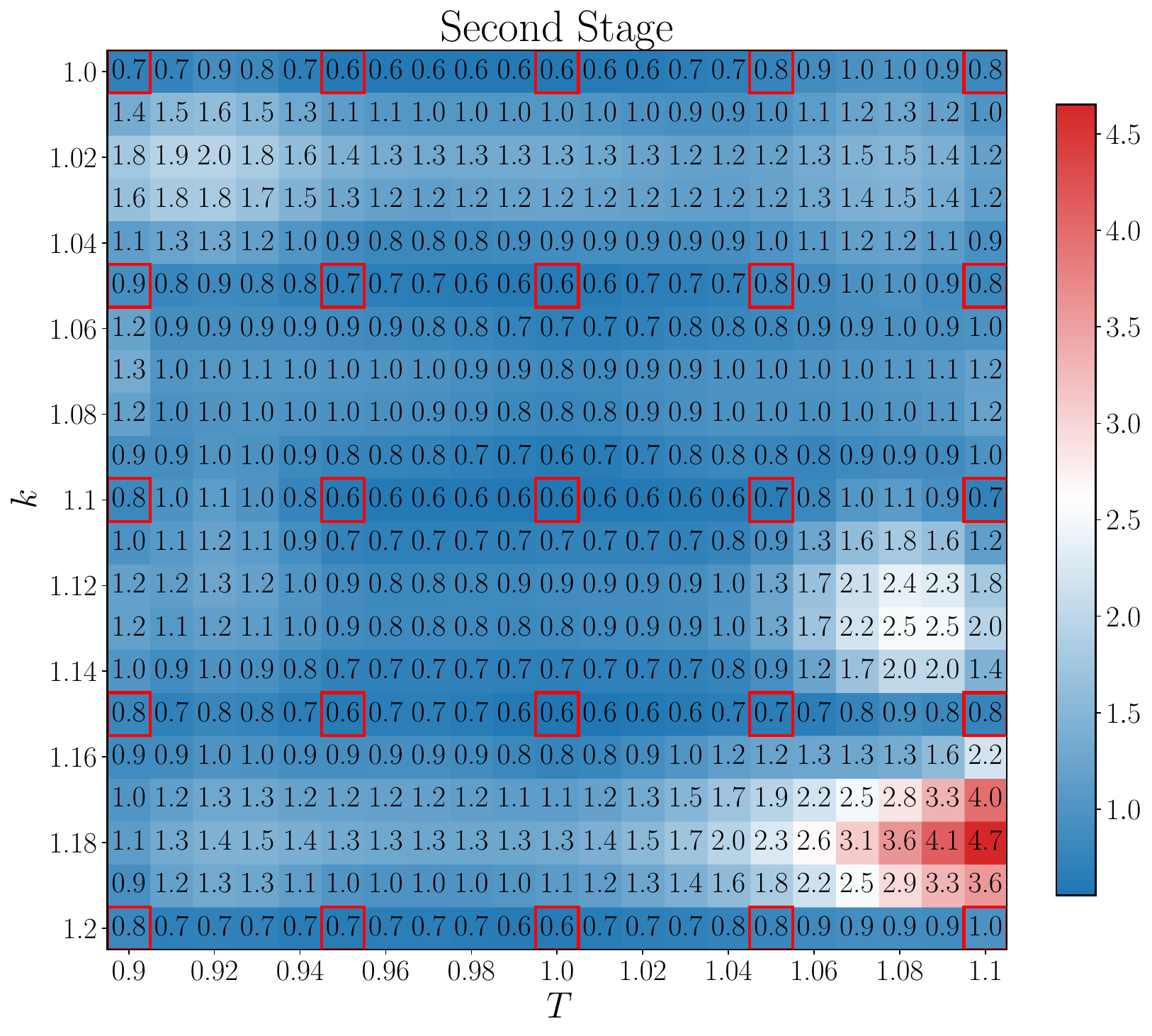}
  \caption{Maximum relative errors when applying mLaSDI with two stages to 1D-1V Vlasov equation with architecture 4096-500-5, where each stage is trained for 25,000 iterations. Red boxes indicate training parameters and the rest are predicted. We see that the second stage of training significantly lowers the maximum relative error achieved compared to only one stage of training.}
  \label{fig:1d1v_comp}
\end{figure}

Figure~\ref{fig:1d1v_comp} shows the error distribution across the parameter space for a representative mLaSDI model with architecture 4096-500-5.
After the first stage, errors range from approximately 7--9\% throughout the domain.
The second stage reduces errors substantially, achieving below 2\% for most parameters and below 1\% near the training points.
%

\begin{figure}
  \centering
      \includegraphics[width=\linewidth]{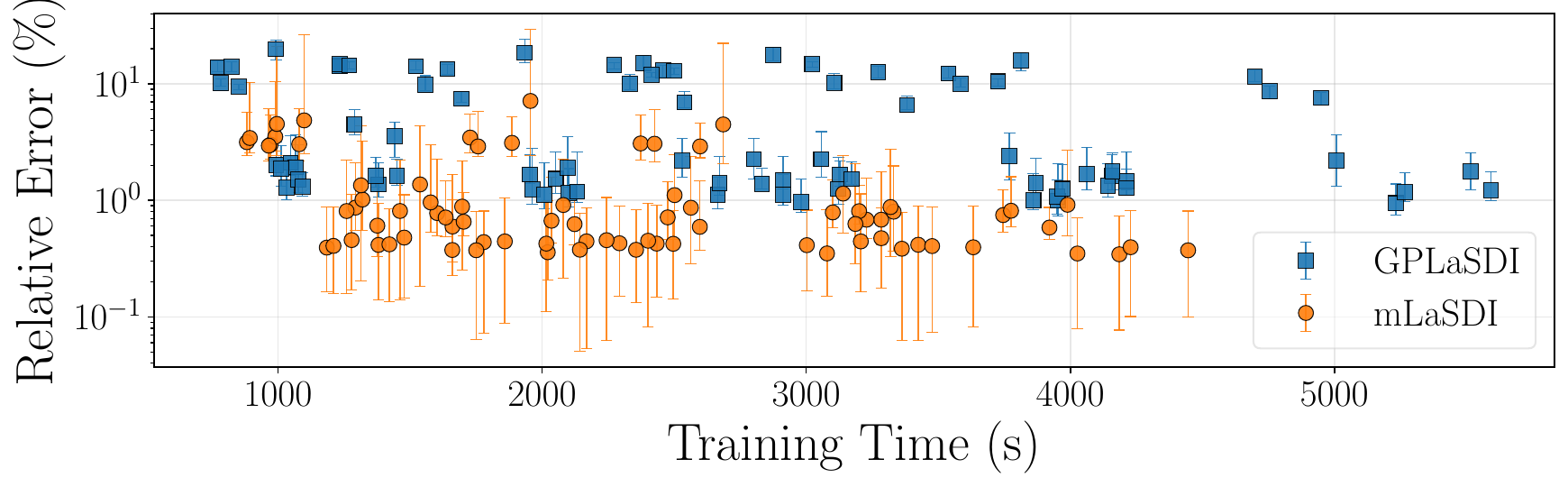}
  \caption{Applying mLaSDI and GPLaSDI to 1D-1V Vlasov equation using a wide range of architectures in Table~\ref{tab:1d1v}. Error bars indicate the maximum and minimum relative errors throughout the parameter space, and markers indicate the median relative error (using the error metric~\eqref{eq:maxerrs}). 
  }
  \label{fig:1d1v_std}
\end{figure}

Figure \ref{fig:1d1v_std} shows the errors from each model along with training time.
The reported training times for mLaSDI are the total time to train both the first and second stages.
We see that mLaSDI consistently produces lower maximum, mean, and minimum relative errors than GPLaSDI models despite using fewer model parameters.
For some models, introduction of a second stage also allows us to achieve maximum relative errors below 1\%, which we never achieve by training only one stage of GPLaSDI.
Aggregating across all configurations, mLaSDI reduces the median error by factors of 2.54, 3.24, and 5.91 for the maximum, mean, and minimum errors, respectively compared to GPLaSDI models trained for the same number of iterations.
%
We also see that mLaSDI has much greater variance between the maximum and minimum relative errors compared to GPLaSDI.
This is expected, as the second stage in mLaSDI minimizes the reconstruction error on training data and so reconstruction error at these points will be much lower than GPLaSDI or the first stage.

%

%

\section{Discussion}
\label{sec:disc}
We have introduced multi-stage Latent Space Dynamics Identification (mLaSDI), a framework that extends LaSDI variants by sequentially training decoders to correct residual errors from earlier stages.
By decoupling reconstruction accuracy from the dynamics-constrained first stage, mLaSDI overcomes the fundamental tension between interpretable latent dynamics and representational capacity that limits standard LaSDI approaches.
Numerical experiments on a multiscale oscillating system, unsteady wake flow past a cylinder, and the 1D-1V Vlasov equation demonstrate that mLaSDI consistently achieves higher training and prediction accuracy over GPLaSDI.
These improvements stem from the multi-stage architecture rather than increased model capacity, and the second-stage decoder efficiently captures high-frequency residuals that the dynamics-constrained first stage cannot represent.

Our results suggest that LaSDI variants face a tension between interpretability of dynamics in the latent space and representational capacity. 
Smooth latent dynamics are desirable for interpretability and stability of our models, but limit accurate reconstruction of high-frequency data. 
Multi-stage training offers a resolution to this problem.
The first stage captures interpretable, low-frequency dynamics, while subsequent stages recover fine-scale features without sacrificing interpretability of the latent space.

Several limitations and interesting aspects of mLaSDI must still be explored.
One possible approach is integrating weak SINDy \cite{Messenger2021} rather than strong SINDy, which has been shown to improve prediction accuracy even without noisy data \cite{He2025,Tran2024}.
It remains unclear how to apply mLaSDI without overfitting noisy data, and mLaSDI may overfit noise during higher stages.
Another consideration is that we restricted our attention to linear latent dynamics for stability reasons, but richer dynamics could improve accuracy if instabilities can be controlled.
Additionally, the choice of two stages was sufficient for all examples considered, but systematic guidelines for selecting the number of stages and their relative training budgets remain to be developed.

{
An important direction for future work is the incorporation of additional physical structure into LaSDI and mLaSDI. 
In this work, we evaluate the models only in the interpolation regime and do not assess extrapolation outside the training time horizon or parameter domain. 
As with many purely data-driven, non-intrusive reduced-order models, we expect prediction accuracy to rapidly deteriorate outside the training regime. 
Improving robustness in such settings may require incorporating physics directly into the autoencoder architecture or imposing additional physics-informed constraints on the learned latent-space ODEs.}
These extensions offer exciting opportunities for further advancing the mLaSDI framework.


\section*{Declaration of competing interests}
The authors declare that they have no known competing financial interests or personal relationships that could have appeared to influence the work reported in this paper.

\section*{Acknowledgements}
This work was partially supported by the Lawrence Livermore National Laboratory (LLNL) under Project No. 50284.
Robert Stephany was supported by the Sydney Fernbach Postdoctoral Fellowship under LDRD number 25-ERD-
049.
Y.\ Choi was also supported for this work by the U.S. Department of Energy, Office of Science, Office of Advanced Scientific Computing Research, as part of the CHaRMNET Mathematical Multifaceted Integrated Capability Center (MMICC) program, under Award Number DE-SC0023164.
Livermore National Laboratory is operated by Lawrence Livermore National Security, LLC, for the U.S. Department of Energy, National Nuclear Security Administration under Contract DE-AC52-07NA27344. 
LLNL document release number: LLNL-JRNL-2014430.

\bibliography{references.bib}

\newpage
\appendix

{
\section{Error propagation with LaSDI}
\label{app:Error_Bound}
\subsection{Proof of proposition \ref{prop:errorbound}}
\label{app:prop1_proof}
\begin{proof}
Define the latent trajectory error
\begin{equation}
    \pmb \eps_z(t) := \vc z(t) - \hat{\vc z}(t).
\end{equation}
Differentiating and using \eqref{eq:latent_approx_ode}, we obtain
\begin{align}
    \dot{\pmb \eps}_z(t) &= \dot{\vc z}(t) - \hat{ \vc f }(\hat{\vc z}(t)) = \vc r(t) + \hat{ \vc f }(\vc z(t)) - \hat{ \vc f }(\hat{\vc z}(t)).
\end{align}
Taking norms and using the Lipschitz continuity of $\hat{ \vc f }$,
\begin{equation}
    \|\dot{\pmb \eps}_z(t)\| \leq \|\vc r(t)\| + L_f \|\pmb \eps_z(t)\|.
\end{equation}
Applying the Cauchy-Schwarz inequality, we have
\begin{equation}
    \frac{d}{dt}\|\pmb \eps_z(t)\| \leq \| \dot{\pmb \eps}_z \| \leq L_f \|\pmb \eps_z(t)\| + \|\vc r(t)\|.
    \label{eq:ez_diff_ineq}
\end{equation}
Since $ \hat{\vc z}(0) = \vc z(0) $, we have $ \pmb \eps_z(0) = 0 $. Applying
Gronwall's inequality to \eqref{eq:ez_diff_ineq} yields
\begin{equation}
    \|\pmb \eps_z(t)\| \leq \int_0^t e^{L_f(t-s)} \|\vc r(s)\| \ ds \leq \eps_{\mathrm{Dyn}}\int_0^t e^{L_f(t-s)} \ ds .
    \label{eq:ez_gronwall}
\end{equation}

We now estimate the state-space error from our SINDy trajectories,
\begin{align}
    \|\vc u(t) - \hat{\vc u}(t)\| &= \|\vc u(t) - D(\hat{\vc z}(t))\| \\
    &\leq \|\vc u(t) - D(\vc z(t))\| + \|D(\vc z(t)) - D(\hat{\vc z}(t))\| \\
    &\leq \eps_{\mathrm{AE}} + L_D \|\pmb \eps_z(t)\|.
\end{align}
Substituting \eqref{eq:ez_gronwall} gives \eqref{eq:lasdi_bound_general}. The
bound \eqref{eq:lasdi_bound_sup} follows from the estimate $\|\vc r(s)\| \leq \eps_{\mathrm{Dyn}}$
for all $ s \in [0,T]$.
\end{proof}

\subsection{Proof of proposition \ref{prop:stage_mono}}

\begin{proof}
  \label{app:prop2_proof}

  For $k \ge 2$, by definition of $\vc R_k$,
\begin{equation}
\vc R_k = \vc U - \tilde{\vc U}_1 - \sum_{i=2}^{k} \eps_{i-1}\tilde{\vc U}_i = \vc R_{k-1} - \eps_{k-1}G_{\mathrm{dec},k}(\hat{\vc Z};\theta_{\mathrm{dec},k}^\star).
\end{equation}
Therefore,
\begin{equation}
\|\vc R_k\|^2 = \eps_{k-1}^2 \left\| \eps_{k-1}^{-1}\vc R_{k-1} - G_{\mathrm{dec},k}(\hat{\vc Z};\theta_{\mathrm{dec},k}^\star) \right\|^2 = \eps_{k-1}^2 \mathcal L_{\mathrm{dec},k}(\theta_{\mathrm{dec},k}^\star).
\end{equation}
Since the zero map belongs to the admissible class and
$\theta_{\mathrm{dec},k}^\star$ is a global minimizer,
\begin{equation}
\mathcal L_{\mathrm{dec},k}(\theta_{\mathrm{dec},k}^\star) \le \mathcal L_{\mathrm{dec},k}(0) = \left\| \eps_{k-1}^{-1} \vc R_{k-1} \right\|^2.
\end{equation}
Hence,
\begin{equation}
\|\vc R_k\|^2 \le \eps_{k-1}^2 \left\| \eps_{k-1}^{-1}\vc R_{k-1} \right\|^2 = \|\vc R_{k-1}\|^2.
\end{equation}
Taking square roots gives $\|\vc R_k\| \le \|\vc R_{k-1}\|$. The strict case
follows immediately if
$\mathcal L_{\mathrm{dec},k}(\theta_{\mathrm{dec},k}^\star) < \mathcal L_{\mathrm{dec},k}(0)$.
\end{proof}

}

\section{Testing scaling factors}
\label{app:kappatesting}

\begin{figure}
  \centering
      \includegraphics[width=\linewidth]{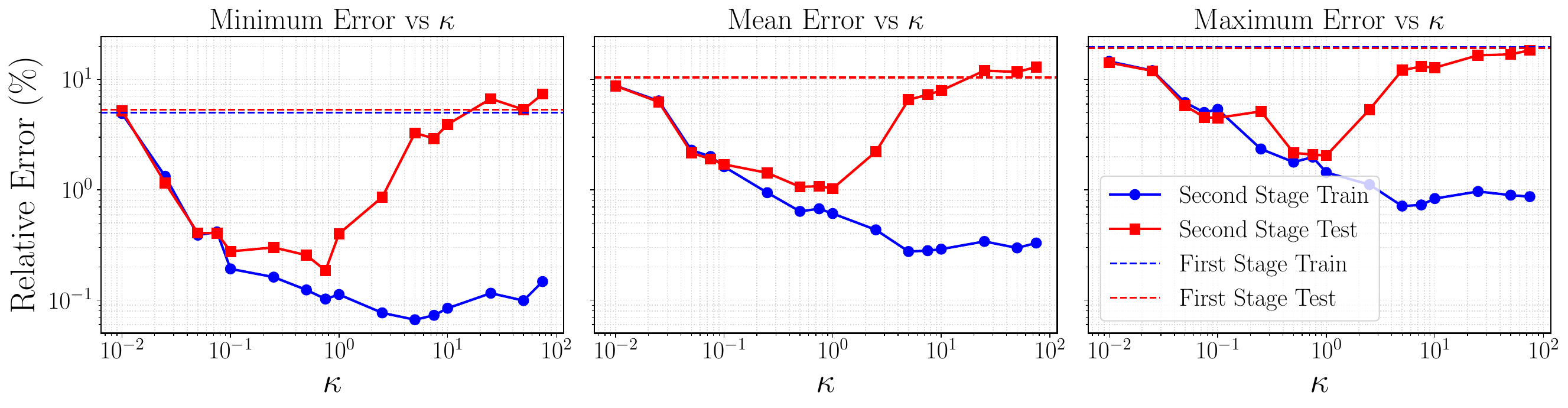} 
  \caption{Effect of scaling factor $\kappa$ on mLaSDI performance for the multiscale problem~\eqref{eq:multistage_func}. Each stage is trained for 10,000 epochs. Test error is minimized near $\kappa = 1$, where larger values lead to overfitting, while smaller values slow convergence.}
  \label{fig:kapptest}
\end{figure}

The multistage networks of \citet{Wang2024} scale weight initialization by a factor $\kappa > 0$ to accelerate residual learning in later stages.
We examine how scaling the standard deviation of the weight distribution for the first layer (with sinusoidal activation) of the second-stage decoder.

We initialize scaled layers using the Xavier normal distribution \cite{glorot2010understanding}.
For a layer mapping $\mathbb{R}^{n_{\text{in}}}$ to $\mathbb{R}^{n_{\text{out}}}$, weights are sampled from $N\left( 0, G\sqrt{\tfrac{1}{n_{\text{in}} + n_{\text{out}}}} \right)$, where $G > 0$ is the gain.
For fully connected networks, we typically use $G = 1$, though \cite{Wang2024} suggests $G = \kappa$ for the second stage.

Since samples from $N(\mu, \sigma)$ scaled by $\eta > 0$ are distributed as $N(\eta \mu, \eta \sigma)$, using Xavier initialization with gain $\kappa$ is equivalent to using gain $1$ and then scaling by $\kappa$.
We introduce $\kappa$ into mLaSDI by defining the decoder's first layer $L_1 : \mathbb{R}^L \to \mathbb{R}^{N_d(1)}$ as 
\begin{equation*}
    L_1(z) = \sin\left( \kappa \left( W \mathbf z + \mathbf b \right) \right),
\end{equation*}
where $L \in \mathbb{N}$ is the latent dimension and $N_d(1)$ is the first hidden layer width.
We initialize $W \in \mathbb{R}^{L \times N_{\text{dec}}(1)}$ with samples from $N\left( 0, \sqrt{\tfrac{1}{L + N_{\text{dec}}(1)}} \right)$ and set $\mathbf b = \mathbf 0$.
This is equivalent to sampling initial weights from $N\left( 0, \kappa \sqrt{\tfrac{1}{L + N_{\text{dec}}(1)}} \right)$ but allows $\kappa$ to be trainable.

Figure \ref{fig:kapptest} shows how $\kappa$ (as a fixed parameter) impacts mLaSDI's reconstruction and prediction accuracy across a wide range for the multiscale oscillating system of Section \ref{subsec:Multiscal}.
Training error continues to decrease until for $\kappa$ up to approximately 10. 
However, there is a clear minimum prediction error when setting $\kappa = 1$. 
For $\kappa > 1$, the model overfits the training data, and for $\kappa < 1$ learning slows and longer training is required.
Based on this empirical evidence, we choose to use $\kappa=1$ for all of our numerical experiments.

{
\section{Effect of additional stages}
\label{Appendix:Multi_Stage}

Throughout the main paper, we focus on the case of two stages, which already provides substantial gains in both reconstruction and prediction accuracy.
A natural question is whether continuing to add stages yields further improvement, or whether doing so simply fits increasingly small residuals without improving (and potentially harming) generalization.
In this appendix, we examine that question for the multiscale oscillating system of Section~\ref{subsec:Multiscal}.

As shown in the main text, standard GPLaSDI performs poorly on this problem: it captures the dominant low-frequency envelope in~\eqref{eq:multistage_func}, but largely misses the small high-frequency oscillations.
By contrast, two-stage mLaSDI resolves both components accurately, and its predictions visually match the governing system much more closely.
This makes the multiscale example a useful testbed for studying whether a \emph{third} stage provides any meaningful benefit once the dominant residual has already been removed.

For these experiments, we use the same first-stage autoencoder architecture as in Section~\ref{subsec:Multiscal}, namely 600-200-20-10.
We also fix $\kappa = 1.0$, use the same loss weights $(\beta_1,\beta_2) = (10^{-1},10^{-3})$, and use the same learning rate $10^{-3}$ as in the main text.
All three stages are trained for $10{,}000$ epochs.
Thus, relative to the two-stage configuration studied in the main paper, this experiment appends one additional residual-learning stage while leaving all other hyperparameters unchanged.

\begin{figure}
  \centering
      \includegraphics[width=\linewidth]{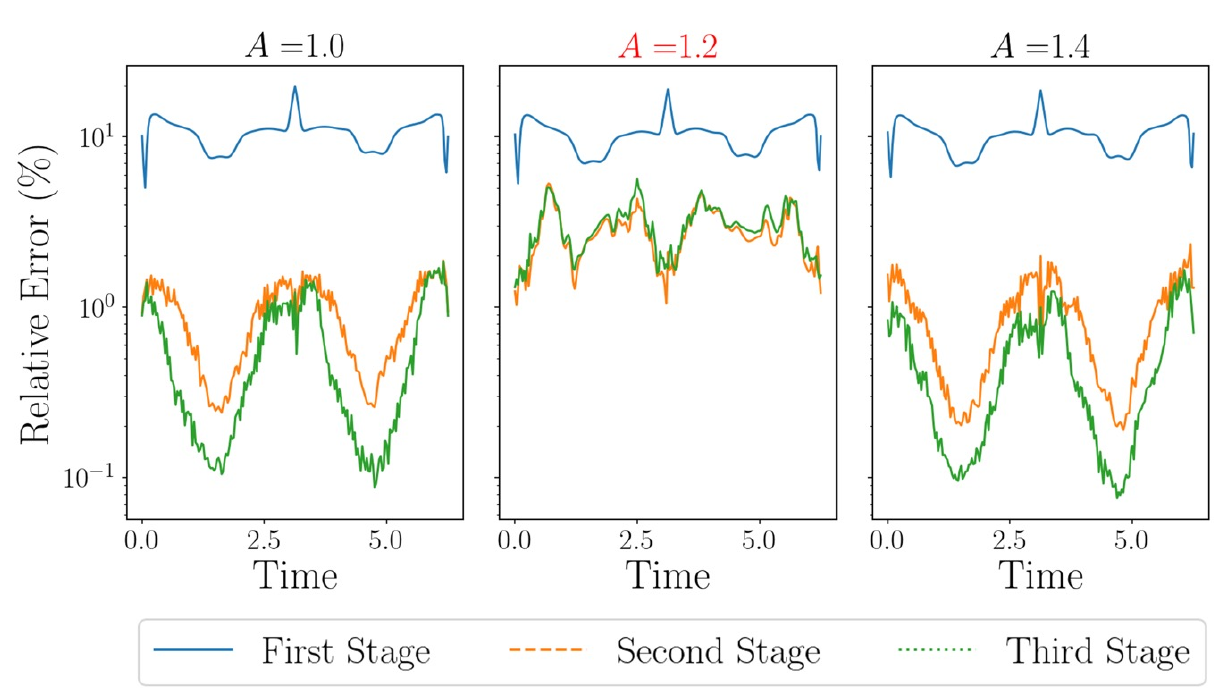}
  \caption{{Relative error for the multiscale oscillating problem~\eqref{eq:multistage_func} using three stages of mLaSDI. Each stage is trained for $10{,}000$ epochs. The third stage slightly reduces the error on the training parameters $A=1.0$ and $A=1.4$, but produces little improvement for the test parameter $A=1.2$, indicating diminishing returns beyond two stages.}}
  \label{fig:three_stages}
\end{figure}

The results are shown in Figure~\ref{fig:three_stages}.
The third stage does reduce the error on the training parameters, indicating that it successfully fits part of the residual left after the second stage.
However, the effect on the test parameter $A=1.2$ is negligible, and in some time intervals the third stage slightly worsens the prediction.
This behavior is consistent with mild overfitting: the third stage can further reduce the residual on the training set, but those corrections do not translate into a meaningful improvement in out-of-sample accuracy.

This outcome is also natural from the perspective of the mLaSDI construction.
After two stages, the remaining residual is already quite small, since the first stage captures the smooth low-frequency behavior and the second stage recovers the dominant missing high-frequency content.
The third stage is therefore trained only on a much smaller residual signal.
Even if this residual is fit accurately on the training set, its contribution to the full approximation
\[
\widetilde{U}_1 + \varepsilon_1 \widetilde{U}_2 + \varepsilon_2 \widetilde{U}_3
\]
is necessarily limited, because it is scaled by the standard deviation of the preceding residual and represents only a small correction to an already accurate prediction.
In other words, once the principal structured error has been removed, subsequent stages have much less signal to learn and are more likely to fit parameter-specific artifacts than broadly transferable features.

We observed the same qualitative trend across a range of nearby hyperparameter choices (not shown): a third stage can slightly improve reconstruction on the training cases, but it does not materially improve prediction accuracy.
For this multiscale problem, these results suggest that two stages are sufficient, and that additional stages beyond apparent convergence yield strongly diminishing returns.
More broadly, they indicate that mLaSDI is not especially sensitive to using ``too many'' stages: adding another stage after convergence does not catastrophically degrade performance, but it also does not appear to offer a practical advantage.
For the examples considered in this work, the main benefit of mLaSDI is obtained once the leading residual has been learned, which in our experiments occurs by the second stage.
}

\section{Instability with second order polynomials}
\label{app:2ndorderSINDy}

In this appendix, we demonstrate that higher-order polynomial terms in the SINDy library can lead to unstable predictions when interpolating for unseen parameters.

Our choice of linear latent dynamics \eqref{eq:lineardynam} is not merely for simplicity, but also for stability. 
Nonlinear terms in the SINDy library can potentially introduce finite-time singularities. 
For instance, the simple scalar ODE $\dot{z} = z^2$ has solution $z(t) = z_0/(1-z_0 t)$ which blows up at finite time $t = z_0^{-1}$. 
In contrast, linear dynamics guarantee existence for any finite time because we have
\begin{equation}
    \dot{\vc z} = A \vc z + \vc b \implies \vc z(t) = \sum_{i=1}^n \alpha_i e^{\lambda_i t} \vc w_i - A^{-1}\vc b,
\end{equation}
where $(\vc w_i, \lambda_i)$ are eigenpairs of $A$ and $\alpha_i$ are coefficients determined by the initial condition.

This stability guarantee is particularly critical in the LaSDI framework for two reasons. 
First, during training, the latent space representation evolves continuously unlike standard SINDy applications to fixed data. 
Nonlinear dynamics can then blow up during training as the autoencoder explores different latent representations. 
Second, we must interpolate SINDy coefficients for unseen parameters. 
Even if dynamics at training parameters $\{\pmb{\mu}^{(i)}\}$ are stable, interpolation can produce coefficient combinations for $\pmb{\mu}^*$ that yield unstable dynamics.
Linear dynamics eliminate the risk of finite-time blowup, and provide greater stability during predictions for unseen input parameters.
This stability constraint motivates our multi-stage approach, rather than pursuing accuracy through richer latent dynamics.

To demonstrate this numerically, we return to the 1D-1V Vlasov equation~\eqref{eq:Vlasov}, applying only one stage of mLaSDI using the same hyperparameters as in Table \ref{tab:1d1v}.
Training is performed on an AMD MI300A GPU via the LLNL Tuolumne cluster.
Here, we expand the SINDy library to allow for pure quadratic terms (e.g., $z_1^2, \ z_2^2$) but exclude the cross terms, such as $z_1z_2$.

\begin{figure}
  \centering
      \includegraphics[width=\linewidth]{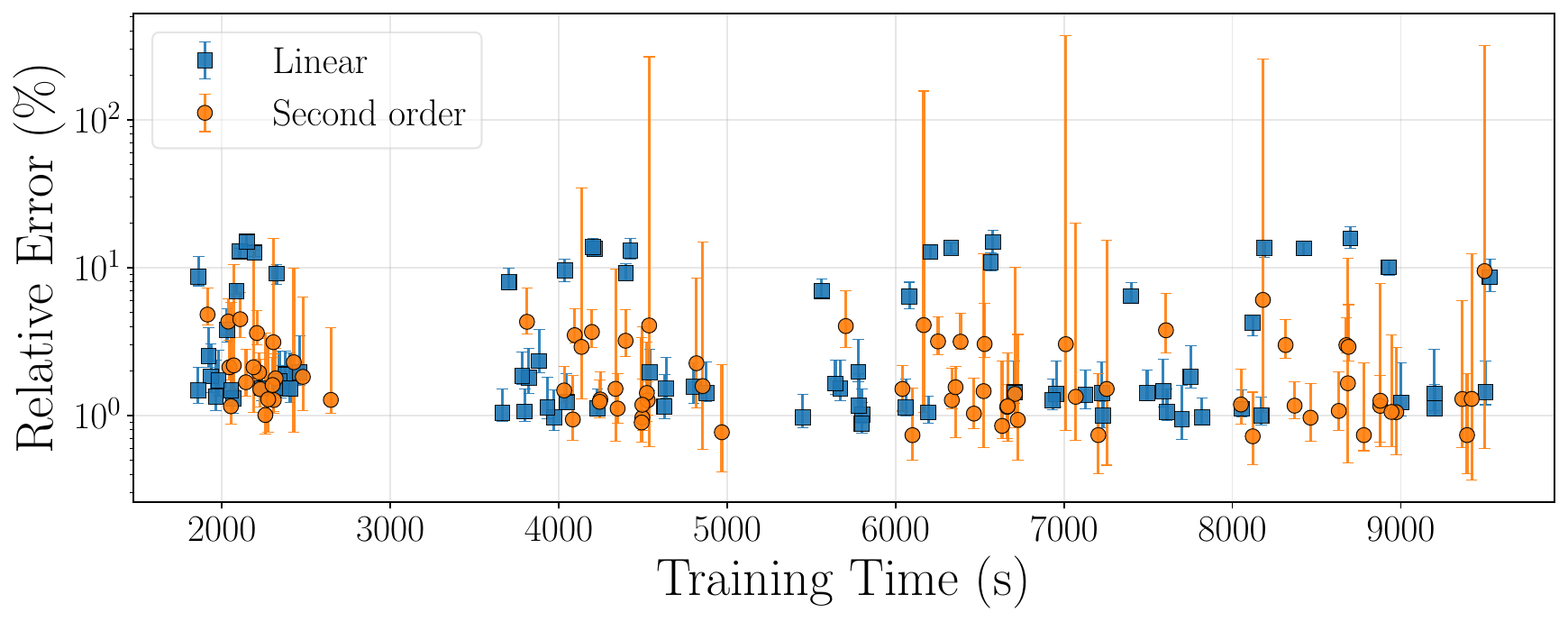}
  \caption{Applying GPLaSDI to 1D-1V Vlasov equation using a wide range of architectures in Table~\ref{tab:1d1v}, with linear and second order dynamics in the SINDy library. Error bars indicate the maximum and minimum relative errors throughout the parameter space, and markers indicate the median relative error.}
  \label{fig:1d1v_quadcomp}
\end{figure}

Figure \ref{fig:1d1v_quadcomp} compares relative errors across the parameter space when using linear and second order terms in the SINDy library. 
The quadratic terms slightly improve median and minimum errors for most cases.
However, unstable interpolations can lead to significantly larger prediction errors from the second order dynamics, resulting in relative errors over 100\% for some predictions.
We avoid higher-order terms in the SINDy library due to marginal accuracy gains combined with the possibility of unstable predictions at unseen parameter values.

\end{document}